\documentclass{article}

\PassOptionsToPackage{numbers, compress}{natbib}
\usepackage[main, final]{neurips_2026}

\usepackage[utf8]{inputenc} 
\usepackage[T1]{fontenc}    
\usepackage{hyperref}       
\usepackage{url}            
\usepackage{booktabs}       
\usepackage{amsfonts}       
\usepackage{nicefrac}       
\usepackage{microtype}      
\usepackage[dvipsnames]{xcolor}         
\usepackage{graphicx} 
\usepackage{amsmath}
\usepackage{algorithm}
\usepackage{algpseudocode}
\usepackage{tabularx}
\usepackage{multirow}
\usepackage{multicol}
\usepackage{dsfont}
\usepackage{makecell}

\newcommand{\freedave}{FreeDave}
\newcommand{\mask}{\textbf{m}}
\newtheorem{theorem}{Theorem}
\newtheorem{lemma}{Lemma}
\newtheorem{definition}{Definition}
\newtheorem{assumption}{Assumption}
\DeclareMathOperator*{\argmax}{arg\,max}

\newcommand{\up}[1]{\textcolor{PineGreen}{(+#1)}}
\newcommand{\speedup}[1]{\textcolor{PineGreen}{(#1$\times$)}}
\newcommand{\down}[1]{\textcolor{BrickRed}{(-#1)}}

\newcolumntype{C}{>{\centering\arraybackslash}X}
\newcolumntype{L}{>{\raggedright\arraybackslash}X}
\newcolumntype{R}{>{\raggedleft\arraybackslash}X}

\title{Free Draft-and-Verification: Toward Lossless Parallel Decoding for Diffusion Large Language Models}

\author{%
  Shutong Wu \\
  Department of Computer Sciences\\
  University of Wisconsin–Madison\\
  \texttt{swu494@wisc.edu} \\
  \And
  Jiawei Zhang \\
  Department of Computer Sciences\\
  University of Wisconsin–Madison\\
  \texttt{jzhang2924@wisc.edu} \\
}

\begin{document}

\maketitle

\begin{abstract}
Diffusion Large Language Models (dLLMs) have emerged as a compelling paradigm for language modeling, offering the unique potential for highly efficient inference via multi-token parallel decoding. However, unlocking this potential remains challenging. High-quality generation typically defaults to one-token-per-step static decoding, while existing parallel algorithms rely on heuristic confidence thresholds, introducing a fragile trade-off between decoding efficiency and generation quality. To overcome this, we introduce Free Draft-and-Verification (FreeDave), a novel training-free and model-free fast decoding algorithm. FreeDave capitalizes on the inherent properties of dLLMs by treating future token predictions as ``free drafts'' and rigorously self-verifying them in subsequent forward passes. Theoretically, FreeDave searches for a shorter decoding path that is guaranteed to reproduce the exact sequence generated by static decoding, achieving algorithmically lossless acceleration. With a sufficiently large draft budget and a verification frontier that does not retreat, it can use the fewest possible model forward passes among feasible paths. Extensive evaluations on math reasoning and code generation benchmarks demonstrate that FreeDave can boost the number of generated tokens per model forward (TPF) by up to 5.48$\times$ and tokens per second (TPS) by up to 3.94$\times$, while robustly maintaining generation quality and avoiding the task-specific performance degradation faced by heuristic methods. Code is available at \url{https://github.com/cychomatica/FreeDave}.
\end{abstract}

\section{Introduction}
\label{sec:intro}
The advent of Large Language Models (LLMs) \cite{comanici2025gemini, liu2024deepseek, yang2025qwen3, gpt5, llama4, claudecode} built on large transformers \cite{vaswani2017attention} and autoregressive (AR) next-token prediction has marked a revolutionary milestone on the way to artificial general intelligence. 
In recent years, LLMs have made significant strides in math reasoning \cite{imani2023mathprompter, luo2025wizardmath, shao2024deepseekmath}, which requires deep logical and symbolic understanding. Similarly, in the domain of software engineering, LLMs have shown promising achievements on code generation \cite{claudecode, roziere2023code, jiang2024survey, guo2024deepseek, codegemma2024}, offering powerful tools to automate development tasks and improve development productivity.
Recently, Diffusion Large Language Models (dLLMs) \cite{nie2025large, ye2025dream, wang2025trado} have emerged as a compelling new paradigm of language modeling. Inspired by their profound success in continuous data domains such as high-fidelity image and audio synthesis \cite{rombach2022high, peebles2023scalable, kong2021diffwave}, diffusion models have been adapted to the discrete domain, especially language modeling \cite{austin2021structured, sahoo2024simple, shi2024simplified, labs2025mercury, song2025seed, arriola2025block}. Due to their bidirectional attention mechanism, dLLMs are more capable
of capturing the connection of context, showing their unique advantages in challenges like the famous ``reversal curse'' \cite{nie2025large} or learning under data-constrained scenarios \cite{ni2025training}. 
Moreover, since dLLMs are trained to simultaneously estimate the marginal distributions for all tokens in a sequence, they possess an inherent capability for efficient multi-token parallel decoding.

However, despite their great potential, the practical application of dLLMs is hampered by the challenge that the high generation quality often requires the number of decoding steps equal to the sequence length, which performs a one-token-per-step decoding (denoted as static decoding). 
Existing parallel decoding algorithms \cite{wu2025fast, yu2025dimple} attempt to bypass this challenge by decoding multiple tokens with confidence higher than a pre-defined threshold. However, such heuristic-based thresholding inherently introduces a fragile trade-off between efficiency and quality. As our evaluations reveal, threshold-based methods are prone to overconfidence and can cause noticeable performance degradation and exhibit inconsistency, especially in logic-intensive tasks like math reasoning and code generation. 
To guarantee the performance, a high threshold should be used, which can be too conservative and lead to insufficient acceleration. 
Thus, fully unleashing the great potential of parallel decoding with a rigorous guarantee of generation quality remains a critical challenge for dLLMs. 

\begin{figure}[!t]
    \centering
    \includegraphics[width=1\linewidth]{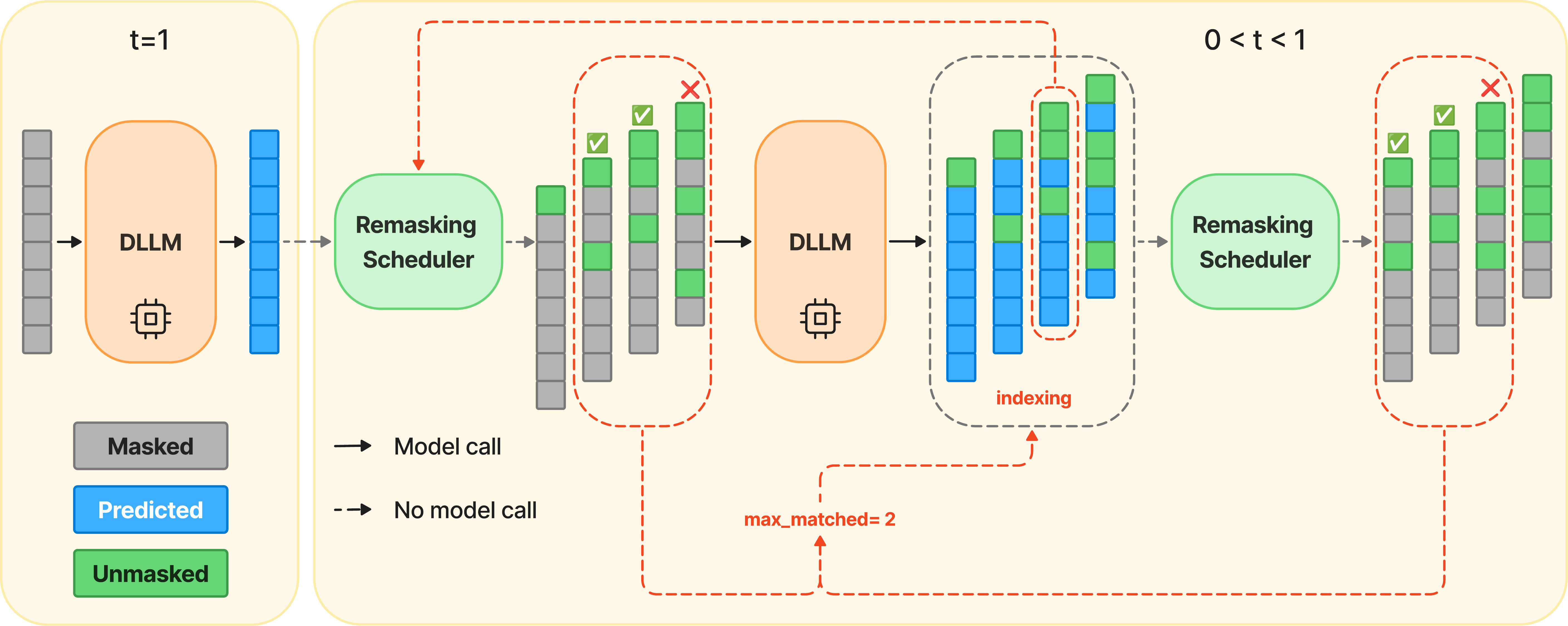}
    \vspace{-10pt}
    \caption{
    The overview of \freedave~decoding. At the current step, the dLLM outputs the estimated distribution of the sequence, from which the remasking scheduler constructs multiple draft candidates. At the subsequent step, the same dLLM processes these candidates in parallel, verifying their validity via sequence matching. Then the resulting output distributions are reused to sample the next set of drafts. Because the logits needed for future drafts are naturally produced as byproducts of the verification step, FreeDave creates a continuous, self-verifying parallel decoding loop that requires no auxiliary models or extra forward passes.
    }
    \vspace{-10pt}
    \label{fig:freedave_pipeline}
\end{figure}

To overcome this challenge, we introduce \textbf{Free} \textbf{D}raft-\textbf{a}nd-\textbf{Ve}rification (\textbf{\freedave}), a novel fast decoding algorithm for dLLMs that achieves lossless parallel decoding. 
Taking advantage of the inherent capability of parallel decoding of dLLMs, and inspired by the idea of draft-and-verification from speculative decoding designed for autoregressive LLMs \cite{leviathan2023fast}, we reveal that: 

\begin{center}
    \emph{Your dLLM is secretly a self-verifiable parallel decoder with almost no extra cost.}
\end{center}
Unlike autoregressive models that require an auxiliary draft model or extra network heads to achieve multi-token parallel decoding via speculative decoding, a dLLM natively estimates the marginal distribution for all tokens in the sequence at every single forward step. FreeDave capitalizes on this unique property: we treat these native future predictions as ``free drafts'' and verify them using the dLLM itself in subsequent standard forward passes. The generation and verification of these candidates are essentially zero-cost byproducts of the normal decoding process. Consequently, FreeDave completely eliminates the need for auxiliary draft models or separate draft-and-verify stages, while seamlessly avoiding the quality-efficiency trade-off that plagues existing parallel algorithms.

In this paper, our contributions are summarized as follows:
\begin{itemize}
\item We introduce \freedave, a training-free and model-free fast decoding algorithm tailored for dLLMs that substantially compresses the total number of decoding steps and accelerates inference without sacrificing generation quality.
\item Theoretically, we provide a guarantee that FreeDave achieves algorithmically lossless decoding.
We prove that FreeDave flawlessly reproduces the exact sequence generated by one-token-per-step static decoding, and can use the fewest possible model forward passes among feasible paths when the verification frontier does not retreat as decoding progresses and the draft budget is sufficiently large.
\item Empirically, we extensively evaluate FreeDave across various dLLMs on diverse benchmarks. Compared to static decoding, \freedave~can boost both the number of generated tokens per model forward (TPF) and tokens per second (TPS) up to over $2\times$, while robustly maintaining task performance where previous threshold-based methods can fail. 
\item Furthermore, \freedave~also serves as a versatile framework capable of pushing heuristic approaches (\emph{e.g.}, threshold-based parallel decoding) to their efficiency limits with preserved generation quality. With a threshold-based scheduler incorporated, compared to static decoding, \freedave~can boost TPF up to $5.48\times$ and TPS up to $3.94\times$, while maintaining task performance.
\end{itemize}

\section{Related Work}
\label{sec:related_work}

Diffusion Large Language Models (dLLMs) offer a compelling alternative to autoregressive (AR) models. Built upon discrete absorbing diffusion \cite{austin2021structured} and Masked Language Modeling \cite{sahoo2024simple, shi2024simplified}, scaled DLLMs \cite{nie2025large, labs2025mercury, ye2025dream, wang2025trado} have closed the performance gap with AR baselines. Their bidirectional attention overcomes AR limitations like the ``reversal curse'' \cite{nie2025scaling} and improves data efficiency \cite{zhang2025survey, ni2025training}. However, this bidirectionality fundamentally complicates standard KV caching, creating a severe bottleneck in inference speed.

To mitigate this bottleneck, specialized caching strategies have been proposed, such as semi-autoregressive Block Diffusion \cite{arriola2025block}, feature-redundant dLLM-Cache \cite{liu2025dllm}, and approximate caching in Fast-dLLM \cite{wu2025fast}. Beyond caching, parallel decoding methods aim to accelerate generation by simultaneously unmasking multiple tokens. Fast-dLLM \cite{wu2025fast} and Dimple \cite{yu2025dimple} rely on heuristically predefined confidence thresholds to accept tokens. However, these heuristic-based methods suffer from a fragile efficiency-quality trade-off, often leading to suboptimal decoding efficiency, or non-negligible performance degradation compared to static decoding \cite{wang2025trado}. Attempts to adaptively mix dLLM and AR distributions \cite{israel2025accelerating} still struggles to eliminate this trade-off completely.

Some recent works also introduce the ``draft-and-verify'' concept, which originates from speculative decoding~\cite{leviathan2023fast, chen2023accelerating, spector2023accelerating, zhang2024draft, cai2024medusa}, to the context of dLLMs, but they tackle fundamentally different research problems. For instance, TiDAR \cite{liu2025tidar} trains a hybrid model that switches between a diffusion mode for drafting and an AR mode for verification. DFlash \cite{chen2026dflash}, on the other hand, trains a dLLM specifically to act as a parallel-decoding draft model to accelerate a target AR model under the traditional speculative decoding framework. However, our work focuses exclusively on accelerating pre-trained dLLMs and introduces a fully plug-and-play, training-free, and model-free algorithm, elegantly treating the pre-trained dLLM as its own drafter and verifier to unlock lossless parallel decoding.

\section{Methodology}
\label{method}
\subsection{Preliminaries}
Like diffusion models on continuous spaces, dLLMs model the transition from a data distribution to a known prior noise distribution on discrete spaces by a forward noising process, and generate data from noise by the induced reverse denoising process. The forward process diffuses the data distribution toward a uniform distribution over the discrete space, or an absorbing state, and the latter is usually adopted for language modeling \cite{sahoo2024simple, shi2024simplified, nie2025large, ye2025dream, wang2025trado}. For the absorbing diffusion, the forward process progressively replaces each unmasked token in the original sequence $x_0$ independently with a special mask token $\mask \in V$ with an increasing probability \emph{w.r.t.} a time level $t$ evolving from 0 to 1.
Once a token is masked, it will remain masked till $t=1$. Eventually, at $t=1$, all the tokens in the sequence will be $\mask$. This forward process with the absorbing state $\mask$ can be formulated as

\begin{equation}
    \label{eq:dllm_forward}
    q(x_t|x_0) = \prod_{i=1}^L q(x_t^i |x_0^i) = \prod_{i=1}^L \text{Cat}\left(x_t^i; \alpha_t \mathbf{e}_{x_0^i} + (1 - \alpha_t) \mathbf{e}_\mask\right)
\end{equation}

where $x_t \in V^{L}$ is the transferred sequence at the time level $t$, $x_t^i \in V$ represents the $i$-th token of $x_t$, $\alpha_t \in \mathbb{R}$ is the noising schedule monotonically decreasing \emph{w.r.t.} $t \in [0, 1]$ and satisfies $a_0=1, a_1=0$, $\mathbf{e}_j \in \{0,1\}^{\vert V \vert}$ is the one-hot vector of interger $j$ over the vocabulary, and Cat represents the categorical distribution over the vocabulary $V$. 

The reverse process is induced to progressively recover the original sequence at $t = 0$ from an all-mask sequence at $t=1$. Specifically, the reverse process transfer the sequence $x_t$ at a time level $t$ conditioned on $x_0$ to the sequence $x_s$ at an earlier time level $s$ (where $0 \leq s < t \leq 1$) by the product of the marginal distributions of each token $q(x_s|x_t, x_0) = \prod_{i=1}^L q(x^i_s|x^i_t, x^i_0)$, where the marginal distribution of the $i$-th token can be formulated as
\begin{gather}
    q(x^i_s|x^i_t, x^i_0) 
    = \begin{cases}
        \text{Cat}(x^i_s; \mathbf{e}_{x^i_t}) & \text{if } x^i_t \neq \mask \\
        \text{Cat}(x^i_s; \frac{(1-\alpha_s)\mathbf{e}_\mask + (\alpha_s - \alpha_t)\mathbf{e}_{x^i_0}}{1 - \alpha_t}) & \text{if } x^i_t = \mask \\
    \end{cases}
\end{gather}
and once a token is unmasked, it will remain unchanged till $t=0$. The challenge is that $x_0$ is agnostic at $t > 0$. As a solution, a parameterized model $f_\theta(\cdot, \cdot): V^{L} \times \mathbb{R} \rightarrow \mathbb{R}^{L \times \vert V \vert}$ is trained to estimate the marginal distributions of each token at $t=0$ by $f^i_{\theta}(x_t, t)$. Given the sequence $x_t$ at time level $t$, the the marginal distribution of the $i$-th token can be calculated by
\begin{gather}
    \label{eq:dllm_decoding_1}
    p_\theta(x^i_s|x^i_t) = 
    \begin{cases}
        \text{Cat}(x^i_s; \mathbf{e}_{x^i_t}) & \text{if } x^i_t \neq \mask \\
        \text{Cat}(x^i_s; \frac{(1-\alpha_s)\mathbf{e}_\mask + (\alpha_s - \alpha_t)f^i_\theta(x_t, t)}{1 - \alpha_t}) & \text{if } x^i_t = \mask \\
    \end{cases}
\end{gather}
such that we can predict the sequence at $s < t$ by $p_\theta(x_s|x_t) = \prod_{i=1}^L p_\theta(x^i_s|x^i_t)$.
Empirically, for better generation quality, the transition from $x_t$ to $x_s$ is not by directly sampling $x_s$ from $p_\theta(x_s|x_t)$. Instead, we have a remasking scheduler $g$ (usually rule-based and unparameterized) 
to decide which $\mask$ tokens to be transferred to unmasked tokens at each time level. The remasking scheduler $g$ takes as input a source time level $t$, a target time level $s$, and the sequence $x_t$, the estimated distribution of $x_0$ at the source time level $t$ (\emph{i.e.} $f_\theta(x_t, t)$ here), and returns a decision set $\mathcal{I}_{t \rightarrow s}$, which is defined as $\mathcal{I}_{t \rightarrow s} = \{(a_1, z_1), \dots, ( a_n, z_n)\}$, where in each tuple $a_j$ is the transferred index and $z_j \sim \text{Cat}(f^{a_j}_\theta(x_t, t))$ is the transferring target token sampled from the estimated distribution. 

Generally, for a high generation quality, the total number of decoding steps is set equal to the response length, and at each step, among all the masked positions, the one where the predicted token has the highest confidence is unmasked. 
We denote this one-token-per-step greedy decoding strategy as static decoding. Unless specified, in this paper, we focus on the static remasking scheduler $g_{\text{static}}$. Details of the static decoding algorithm can be found in Algorithm~\ref{alg:mdlm_original}.

\subsection{Decoding Algorithm: Free Draft-and-Verification}
\label{sec:theory}
\begin{definition}[Decoding Path and Oracle Path]
\label{def:decoding_path}
For the reverse process of dLLM $f_\theta$, a decoding path of $N$ steps following a time schedule array $[t_0, \dots, t_N | 1 = t_0 > t_1 > \dots > t_{N-1} > t_N = 0]$ is defined as $\mathbf{p} = [\mathcal{I}_{t_0 \rightarrow t_1}, \dots, \mathcal{I}_{t_{N-1} \rightarrow t_N}]$, where $\mathcal{I}_{t_i \rightarrow t_{i+1}}$ is the decision set given at timestep $t_i$. 
Given a remasking scheduler $g$ and a predefined time schedule array $\mathbf{T} = [t_0, \dots, t_N | 1 = t_0 > t_1 > \dots > t_{N-1} > t_N = 0]$, we define the oracle path as $\mathbf{p}_{\text{oracle}} = [\mathcal{I}_{t_0 \rightarrow t_1}, \dots, \mathcal{I}_{t_{N-1} \rightarrow t_N}]$, where $\mathcal{I}_{t_i \rightarrow t_{i+1}} = g(x_{t_i}, f_\theta(x_{t_i}, t_i), t_i, t_{i+1})$ is the decision set returned by the remasking scheduler $g$ at time step $t_i$. 
\end{definition}

\begin{definition}[Feasible Path and Optimal Path]
\label{def:optimal_decoding_path}
For the reverse process of dLLM $f_\theta$, given a remasking scheduler $g$ and a predefined time schedule array $\mathbf{T} = [t_0, \dots, t_N | 1 = t_0 > t_1 > \dots > t_{N-1} > t_N = 0]$, if we can find a $M$-step subarray $[t_{a_0}, \dots, t_{a_M} | 0 = a_0 < a_1 < \dots < a_{M-1} < a_M = N, a_i \in \{0, \dots, N\}]$, such that 
\begin{equation}
\begin{split}
    &~\mathcal{I}_{t_{a_i} \rightarrow t_{a_{i+1}}} = ~g(x_{t_{a_i}}, f_\theta(x_{t_{a_i}}, t_{a_i}), t_{a_i}, t_{a_{i+1}}) \\
    =&~\bigcup_{j=0}^{a_{i+1} - a_i - 1} g(x_{t_{a_i + j}}, f_\theta(x_{t_{a_i + j}}, t_{a_i + j}), t_{a_i + j}, t_{a_i + j + 1}) \\
    =&~\bigcup_{j=0}^{a_{i+1} - a_i - 1} \mathcal{I}_{t_{a_i + j} \rightarrow t_{a_i + j + 1}} 
    ,~\forall i \in \{0, \dots, M-1\},
\end{split}
\end{equation}
then the decoding path $\mathbf{p} = [\mathcal{I}_{t_{a_0} \rightarrow t_{a_1}}, \dots, \mathcal{I}_{t_{a_{M-1}} \rightarrow t_{a_M}}]$ reproduces the oracle sequence. For our theoretical analysis, we additionally require each merged interval to admit a consecutive one-step verification chain, as specified in Definition~\ref{def:ordered_space} in Appendix~\ref{apdx:lemma_setting}. A path satisfying both conditions is defined as a feasible path, and $\mathcal{P}(f_\theta, g, \mathbf{T})$ denotes this ordered one-step-verifiable space.
There exists an optimal path $\mathbf{p}^\star =[\mathcal{I}_{t_{a^\star_0} \rightarrow t_{a^\star_1}}, \dots, \mathcal{I}_{t_{a^\star_{M-1}} \rightarrow t_{a^\star_M}}]$ such that $\vert \mathbf{p}^\star \vert \leq \vert \mathbf{p} \vert \leq \vert \mathbf{p}_{\text{oracle}} \vert, \forall \mathbf{p} \in \mathcal{P}(f_\theta, g, \mathbf{T})$, \emph{i.e.} among all feasible paths, the optimal path requires the fewest decoding steps, while the oracle path requires the most decoding steps.
\end{definition}

Crucially, any feasible path $p \in \mathcal{P}(f_\theta, g, T)$ structurally guarantees the reproduction of the exact same output sequence as the oracle path $\mathbf{p}_{\text{oracle}}$. Thus, we formally define decoding along any feasible path as lossless decoding.
Definition~\ref{def:decoding_path} and \ref{def:optimal_decoding_path} indicate the great potential of dLLMs for lossless parallel decoding. 
However, except for the oracle path $\mathbf{p}_{\text{oracle}}$, all of the other lossless feasible paths are agnostic to us. 
We don't know which steps can be merged into one step. Empirically setting a fixed number of transferred tokens or a heuristic confidence threshold for parallel decoding is not guaranteed to get a feasible path that generates the same sequence as decoding via the oracle path, and the experimental results from both~\citet{wang2025trado} and us demonstrate that threshold-based parallel decoding is not always robust and can result in a non-negligible performance degradation. 

Following Appendix~\ref{apdx:lemma_setting}, let $D_{n,r}$ be the nested draft merging the next $r$ oracle transitions from a trusted state $x_{t_n}$, and let $T_{n,r}$ be its one-step target computed in the next batched forward. The first draft $D_{n,1}$ is the one-step oracle successor. For $0\leq n<j\leq N$, we write $F(n,j)=1$ when $T_{n,r}=D_{n,r+1}$ for every $r=1,\dots,j-n-1$. Thus, verification accepts only a consecutive prefix and stops at the first mismatch; a later match beyond that mismatch does not certify a feasible merge.

\begin{theorem}[Verification-based Feasible Path Search]
\label{thrm:feasible_path_search}
In the ordered one-step-verifiable space $\mathcal{P}(f_\theta, g, \mathbf{T})$, let $d$ be a positive integer search budget and $d_n=\min(d,N-n)$. Define the verifier as the length of the longest consecutively verified prefix:
\begin{equation}
    \label{eq:verifier}
    h(n,d)=\max\left\{\ell\in\{1,\dots,d_n\}\;\middle|\;
    T_{n,r}=D_{n,r+1},\ \forall r=1,\dots,\ell-1\right\}.
\end{equation}
For $\ell=1$, the set of verification indices $\{1,\dots,\ell-1\}$ is empty, so the condition is automatically satisfied. This corresponds to accepting the first draft $D_{n,1}$, which is already the one-step oracle successor and requires no target matching. Thus, $\ell=1$ is always admissible, giving $1\leq h(n,d)\leq d_n$. Starting from $a_0=0$ and applying $a_{i+1}=a_i+h(a_i,d)$ until $a_M=N$ yields a feasible path $\mathbf{p} = [\mathcal{I}_{t_{a_0} \rightarrow t_{a_1}}, \dots, \mathcal{I}_{t_{a_{M-1}} \rightarrow t_{a_M}}]$ that reproduces the oracle output.
\end{theorem}

Let $R(n)=\max\{j:F(n,j)=1\}$ be the farthest verifiable endpoint. For the following optimality result, we assume Verification-Frontier Non-Retreat (Assumption~\ref{assump:vfnr}) holds: $F(i,j)\Rightarrow F(k,j)$ for $i<k<j$, or equivalently $R(i+1)\geq R(i)$. Details can be found in Appendix~\ref{apdx:lemma_setting}.

\begin{lemma}[Step Optimality]
\label{lemma:optimal_path_guarantee}
Under Assumption~\ref{assump:vfnr}, the greedy longest-merge path $\mathbf{p}^\star$, defined by $a^\star_0=0$ and $a^\star_{i+1}=R(a^\star_i)$ until $a^\star_M=N$, has the fewest decoding steps among all paths in the ordered one-step-verifiable space $\mathcal{P}(f_\theta,g,\mathbf{T})$. If
$d\geq\Vert\mathbf{p}^\star\Vert_\infty=\max\{a^\star_{i+1}-a^\star_i\mid i=0,\dots,M-1\}$,
then the verifier in Theorem~\ref{thrm:feasible_path_search} recovers this optimal path, with $a^\star_{i+1}=a^\star_i+h(a^\star_i,d)$ for $i=0,\dots,M-1$.
\end{lemma}
\vspace{-5pt}
\paragraph{Proof Sketch}
{
For Theorem~\ref{thrm:feasible_path_search}, $D_{n,1}=x_{t_{n+1}}$. If $D_{n,r}=x_{t_{n+r}}$, its one-step target is the next oracle state, so a successful check $T_{n,r}=D_{n,r+1}$ gives $D_{n,r+1}=x_{t_{n+r+1}}$. Induction over the consecutive checks in Eq.~\eqref{eq:verifier} therefore yields
\[
\mathcal{I}_{t_n\rightarrow t_{n+h(n,d)}}
=\bigcup_{r=0}^{h(n,d)-1}\mathcal{I}_{t_{n+r}\rightarrow t_{n+r+1}},
\]
which induces a feasible path by Definition~\ref{def:optimal_decoding_path}. When $d=1$, this reduces to the oracle path.

For Lemma~\ref{lemma:optimal_path_guarantee}, compare the untruncated greedy path with any feasible path $0=b_0<\cdots<b_K=N$. Assume $a^\star_r\geq b_r$, which holds at $r=0$. If $a^\star_r<b_{r+1}$, Assumption~\ref{assump:vfnr} and $F(b_r,b_{r+1})=1$ give $F(a^\star_r,b_{r+1})=1$, so greedy reaches at least $b_{r+1}$ on its next step. Otherwise, it has already reached or passed that endpoint. Induction gives $a^\star_r\geq b_r$ until greedy terminates, proving that it reaches $N$ in at most $K$ steps. A budget $d\geq\Vert\mathbf{p}^\star\Vert_\infty$ leaves every greedy merge untruncated and hence recovers this optimal path. The complete proof is provided in Appendix~\ref{apdx:lemma_proof_main}.
}

\paragraph{Remark}
{
Under the conditions of Lemma~\ref{lemma:optimal_path_guarantee}, a draft budget covering the largest number of oracle transitions merged along the untruncated greedy optimal path suffices to recover that path. In particular, $d\geq N \geq \Vert\mathbf{p}^\star\Vert_\infty$ provides a sufficient condition. Detailed analysis is provided in Appendix~\ref{apdx:lemma_proof}.
}

Based on our theoretical analysis above, we propose Free Draft-and-Verification (\freedave), a fast decoding algorithm that searches for the longest consecutively mergable decoding steps, with algorithmic losslessness guaranteed by Theorem~\ref{thrm:feasible_path_search} and step optimality under the conditions of Lemma~\ref{lemma:optimal_path_guarantee}.
Specifically, as detailed in Algorithm~\ref{alg:mdlm_freedave}, \freedave~achieves this by instantiating the theoretical verifier $h(n,d)$ through a parallel sequence matching mechanism (\emph{i.e.}, Line 27–33). At each iteration, FreeDave iteratively utilizes the remasking scheduler to sample a batch of draft candidates with varying unmasking tokens based on the current estimated distribution. At the next step, these candidates are fed into the dLLM in parallel. The subsequent sequential matching process essentially acts as the verifier $h(n,d)$ that identifies the longest consecutive match between the drafted states and the one-step-ahead target states, mathematically satisfying Eq.~\eqref{eq:verifier} and ensuring the accepted path remains strictly in the feasible space (\emph{i.e.}, lossless).

It is worth noting that while \freedave~shares the concept of ``draft-and-verify'' with speculative decoding~\cite{leviathan2023fast, chen2023accelerating, spector2023accelerating, zhang2024draft, cai2024medusa}, their operational paradigms are fundamentally different. Speculative decoding relies on an auxiliary draft model (or specific architectural modifications like auxiliary heads), separating the draft and verification stages into distinct forward passes. 
In contrast, \freedave~introduces zero extra forward passes compared to static decoding. Since a dLLM naturally computes the marginal distribution for all tokens simultaneously, the outputs required to verify the current drafts inherently contain the distributions needed to sample the next group of drafts. In \freedave, drafts are strictly ``free'' byproducts decoded and verified concurrently during the standard forward pass. This single-model, unified-stage approach not only unlocks higher decoding efficiency but also entirely sidesteps the notorious draft-target alignment challenge faced by traditional speculative decoding.

\begin{algorithm}[!ht]
  \small
  \caption{dLLM \freedave~Decoding}
  \label{alg:mdlm_freedave}
  \begin{algorithmic}[1]
    \Require Model $f_\theta$, remasking scheduler $g$, total generation length $L$, total decoding steps $N$, draft steps $d$, time schedule array $\mathbf{T} = \{t_0, \dots, t_N | 1 = t_0 > t_1 > \dots > t_{N-1} > t_N = 0\}$
    \Ensure The generated sequence $x$.
    
    \State $i \gets 0$, $x_{t_0} \gets \mask^{1:L}$, $\tilde{x}_0 \gets f_\theta(x_{t_0}, t_0)$
    \Comment{initialize sequence and its estimated distribution}

    \While{$i < N$}
        \State $\mathbf{X}_{\text{draft}} \gets [~]$, $d_i \gets \min(d, N-i)$
        \For{$k$ in $1:d_i$}
            \State $\mathcal{I}_{t_i \rightarrow t_{i+k}} \gets g(x_{t_i}, \tilde{x}_0, t_i, t_{i+k})$
            \State $x_{\text{draft}} \gets x_{t_i}$
            \For{each $(a_j, z_j)$ in $\mathcal{I}_{t_i \rightarrow t_{i+k}}$}
            \State $x^{a_j}_{\text{draft}} \gets z_j$
                \Comment{replace masked tokens according to decision}
            \EndFor
            \State $\mathbf{X}_{\text{draft}}\text{.append}(x_{\text{draft}})$
        \EndFor
        \If{$i = N-1$}
            \State $x_0 \gets \mathbf{X}_{\text{draft}}[1]$ 
            \State break
        \EndIf
        
        \State $\tilde{\mathbf{X}}_{\text{draft}, 0} \gets f_\theta(\mathbf{X}_{\text{draft}}, t_{i+1:i+d_i})$
            \Comment{parallel verification forward of drafts}
        \State $\mathbf{X}_{\text{target}} \gets [~]$
        \For{$k = 1:d_i$}
            \State $\mathcal{I}_{t_{i+k} \rightarrow t_{i+k+1}} \gets g(\mathbf{X}_{\text{draft}}[k], \tilde{\mathbf{X}}_{\text{draft}, 0}[k], t_{i+k}, t_{i+k+1})$
                \Comment{one more step ahead}
            \State $x_{\text{target}} \gets \mathbf{X}_{\text{draft}}[k]$
            \For{each $(a_j, z_j)$ in $\mathcal{I}_{t_{i+k} \rightarrow t_{i+k+1}}$}
            \State $x^{a_j}_{\text{target}} \gets z_j$
            \EndFor
            \State $\mathbf{X}_{\text{target}}\text{.append}(x_{\text{target}})$
        \EndFor
        \State $m \gets 0$
        \For{$k$ in $1:d_i-1$}
        \Comment{instantiation of verifier $h(n,d)$ from Theorem~\ref{thrm:feasible_path_search}}
            \If {$\mathbf{X}_{\text{draft}}[k+1] = \mathbf{X}_{\text{target}}[k]$}
                \State $m \gets m + 1$
            \Else
                \State break
            \EndIf
        \EndFor
        \State $i \gets i + m + 1$, $x_{t_i} \gets \mathbf{X}_{\text{draft}}[m + 1], \tilde{x}_0 \gets \tilde{\mathbf{X}}_{\text{draft}, 0}[m + 1]$
    \EndWhile
    \State $x \gets x_0$\\
    \Return $x$
  \end{algorithmic}
  \vspace{-1pt}
\end{algorithm}

\subsection{Practical Extensions for Block Diffusion}
While Section~\ref{sec:theory} establishes the theoretical foundation of FreeDave, deploying dLLMs in practice often relies on block diffusion~\cite{arriola2025block} to enable KV caching via inter-block autoregressive and intra-block diffusion decoding. To maximize practical utility, we further introduce two critical extensions.
\paragraph{Block Tree Attention for Memory Efficiency.} A naive implementation of Algorithm~\ref{alg:mdlm_freedave} expands draft candidates along the batch dimension, incurring substantial memory overhead. Inspired by~\citet{cai2024medusa}, we compactly pack all draft candidates into a single sequence and employ a block-level tree attention topology (illustrated in Figure~\ref{fig:block_tree_attn}). This structure prevents redundant KV cache allocation for shared prefixes across different draft candidates and reduce the memory footprint during parallel verification.
\paragraph{Eager Mode for Cross-Block Acceleration.} A fundamental limitation arises when directly applying \freedave~to small block sizes: as the intra-block decoding approaches the block boundary, the number of remaining masked positions naturally becomes smaller than the draft budget $d$. This forces $d$ to be truncated, inevitably bottlenecking parallel efficiency. To circumvent this, we introduce Eager Mode (illustrated in Figure~\ref{fig:eager_mode_detail}). Rather than strictly confining drafts within the current block, Eager Mode permits pre-drafting and pre-verifying tokens across future blocks while maintaining a strict left-to-right block acceptance priority. This not only eliminates boundary-induced truncations but also explicitly enables scaling the draft budget $d$ beyond the block size itself. Crucially, as demonstrated in Appendix~\ref{apdx:eager_mode_details}, Eager Mode is structurally equivalent to a static decoding variant with delayed cache updates, thereby strictly preserving the generation quality while unlocking significantly higher decoding efficiency.



\subsection{Integrating Threshold-Based Parallel Decoding into \freedave}
\label{subsec:threshold_freedave}
To further push the efficiency limits, \freedave~can be integrated with threshold-based parallel decoding. In this setting, \freedave~inherits the generation quality of threshold-based parallel decoding while further compressing the decoding steps. Specifically, at each step, among tokens excluding those already selected by the threshold-based scheduler, \freedave~selects the top $d-1$ most confident tokens and combines them with those threshold-selected tokens to construct $d$ draft candidates. And at the next step, we perform a new threshold-based step on each draft to get target sequences. Then, each target sequence $\mathbf{X}_{\text{target}}[k]$ is compared with drafts $\mathbf{X}_{\text{draft}}[k+1:d]$ for verification. 
We detail the Threshold FreeDave algorithm in Appendix~\ref{apdx:threshold_freedave}.

\section{Experiments}
\label{sec:exp}

\subsection{Experimental Settings}
\label{sec:exp:settings}

Our evaluations focus on math reasoning and code generation tasks. For math reasoning benchmarks, we choose MATH500 \cite{hendrycks2021measuring} and GSM8K \cite{cobbe2021training}.
For code generation benchmarks, we choose MBPP \cite{austin2021program} and HumanEval \cite{chen2021evaluating}. In Appendix~\ref{apdx:qa_eval}, we also provide additional evaluations on Q\&A benchmarks, including GPQA\cite{rein2024gpqa} and TruthfulQA\cite{lin2022truthfulqa}. 
We evaluate \freedave~on full-attention dLLMs including LLaDA-8B-Instruct \cite{nie2025large} and Dream-7B-Instruct \cite{ye2025dream}, and block-attention dLLMs including TraDo-4B-Instruct and TraDo-8B-Instruct \cite{wang2025trado}. We compare our method with \textbf{static decoding}, which follows the oracle path induced from a predefined time schedule, and \textbf{threshold-based parallel decoding} \cite{wu2025fast, yu2025dimple}, which at each step unmasks all positions where the transferring target token confidence exceeds a predefined threshold. 

All the models use block-diffusion generation \cite{arriola2025block} and KV Cache. Specifically, we apply the dual cache~\cite{wu2025fast} to full-attention dLLMs and the prefix cache to block-attention dLLMs. 
We use a maximum generation length of 1024 for all the decoding algorithms, and set the block size to 32 for full-attention models and 4 for block-attention models. Besides, all the decoding algorithms enable an early exit mechanism, where the generation ends if the current block consists of all EOS tokens. 
Regarding the number of draft steps $d$ used for \freedave, we set $d=8$ for the block-attention models, and $d=4$ for the full-attention models. We enable Eager Mode on the block-attention models to unlock higher efficiency under a small block size.

\subsection{Evaluation on Benchmarks}
\label{sec:exp:math}

In this section, we compare the accuracy, tokens generated per forward (\textbf{TPF}), and tokens generated per second (\textbf{TPS}) of different dLLMs on math reasoning and code generation benchmarks. Specifically, TPF and TPS are calculated as the number of generated tokens divided by the Number of Function Evaluations (\textbf{NFE}) and response time, respectively. We do not directly use NFE or response time as our measurement of efficiency, as the generated sequences might have varied lengths due to the early exit mechanism. 

\begin{figure}[!h]
    \centering
    \includegraphics[width=0.99\linewidth]{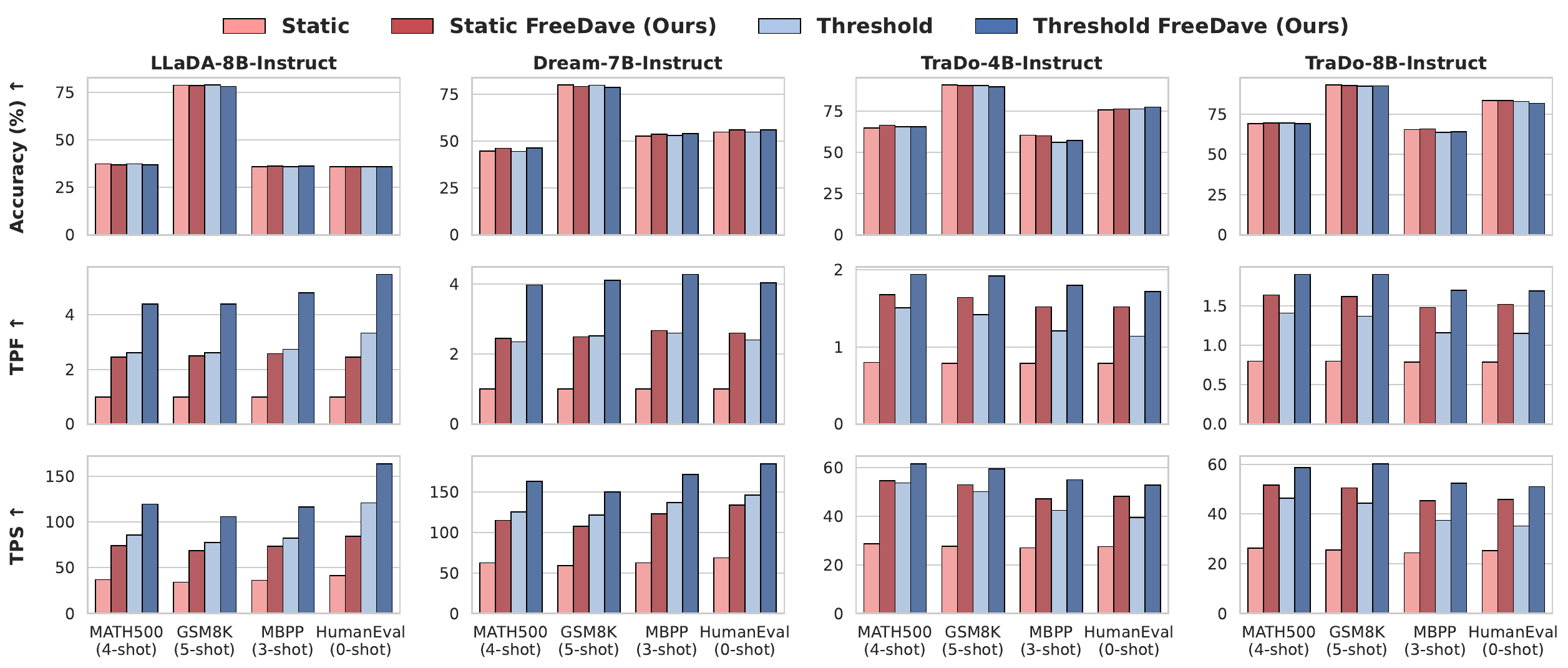}
    \caption{Performance and efficiency comparison of dLLMs with different decoding strategies on different benchmarks. More detailed results can be found in Appendix \ref{apdx:exp_detail}.}
    \label{fig:main_exp}
\end{figure}

\paragraph{Lossless Acceleration via \freedave.} As illustrated in Figure~\ref{fig:main_exp} and detailed in Tables\ref{tab:llada_detail}–\ref{tab:trado8b_detail}, FreeDave demonstrates remarkable inference speedups across all evaluated dLLMs and benchmarks. For block-attention models like TraDo-4B/8B, FreeDave significantly boosts both TPS and TPF (up to over 2x), while algorithmically anchoring the accuracy to the static baseline. The minor empirical fluctuations (within $\pm 1.6\%$) observed are strictly attributed to system-level numerical non-determinism, and we rigorously validate it under a deterministic debug mode, resulting in a 100\% exact match rate (see Section~\ref{sec:non_determinism_analysis}). 

\paragraph{Robustness of \freedave.} While threshold-based parallel decoding~\cite{wu2025fast, yu2025dimple} can sometimes achieve comparable speedups without severe accuracy loss, our evaluations reveal its inherent fragility as a heuristic approach. Because it relies on a predefined confidence scalar rather than rigorous verification, its performance is highly sensitive to the specific model and task. For instance, on logic-intensive tasks like code generation (MBPP), threshold decoding causes a noticeable 4.40\% accuracy drop for TraDo-4B-Instruct. This inconsistency highlights the fundamental dilemma of thresholding: one must meticulously tune the hyperparameter for different tasks and models to avoid arbitrary degradation. Meanwhile, by shifting from a ``guess-and-commit'' heuristic to a ``draft-and-verify'' paradigm, FreeDave provides a structurally robust solution that universally applies to various dLLM architectures without the risk of task-specific performance degradation.

\paragraph{Pushing the Limits: \freedave~with Threshold.} From Figure~\ref{fig:main_exp}, it can be observed that \freedave~built on static decoding is already comparable and even faster than the strong baseline of threshold-based parallel decoding, with a better quality robustness. As introduced in Section~\ref{subsec:threshold_freedave} and detailed in Appendix~\ref{apdx:threshold_freedave}, FreeDave can be further integrated with a threshold-based scheduler to unlock the extreme decoding efficiency. For example, as shown in Figure~\ref{fig:main_exp} and Table~\ref{tab:llada_detail}, compared to static decoding, Threshold FreeDave achieves up to 5.48 TPF (a $5.48\times$ speedup) and 163.75 TPS (a $3.94\times$ speedup) on LLaDA-8B-Instruct, while maintaining the benchmark performance. This demonstrates that our algorithm not only provides a lossless and efficient alternative to static decoding but also serves as a versatile framework capable of pushing heuristic approaches to their efficiency limits with preserved generation quality.

\begin{table*}[!h]
    \centering
    \small
    \begin{tabularx}{\textwidth}{lCCCC} 
    \toprule
    \multirow{2}{*}{\textbf{Model}} & \multicolumn{2}{c}{\textbf{MATH500}} & \multicolumn{2}{c}{\textbf{HumanEval}} \\
    & \textbf{$R_{\text{sequence}}$} & \textbf{$R_{\text{token}}$} & \textbf{$R_{\text{sequence}}$} & \textbf{$R_{\text{token}}$} \\
    \midrule
    LLaDA-8B-Instruct & $500/500$ & $312047/312047$ & $164/164$ & $32977/32977$ \\
    Dream-7B-Instruct & $500/500$ & $178489/178489$ & $164/164$ & $13136/13136$ \\
    TraDo-4B-Instruct & $500/500$ & $290262/290262$ & $164/164$ & $24041/24041$ \\
    TraDo-8B-Instruct & $500/500$ & $296300/296300$ & $164/164$ & $23938/23938$ \\
    \bottomrule
    \end{tabularx}
    \caption{On MATH500 and HumanEval, we evaluate the sequence-level $R_{\text{sequence}}$ and token-level match rate $R_{\text{token}}$ between static decoding and FreeDave decoding under determinism debug mode. }
    \label{tab:determinism_debug}
\end{table*}

\subsection{Disentangling Algorithmic Losslessness from System Numerical Non-determinism}
\label{sec:non_determinism_analysis}

By Theorem \ref{thrm:feasible_path_search}, \freedave~is theoretically guaranteed to traverse a feasible path that reproduces the exact sequence generated via the oracle path of static decoding. However, as observed in Section \ref{sec:exp:math}, the empirical accuracy between static and FreeDave decoding is not perfectly identical, exhibiting minor fluctuations (\emph{e.g.}, within $\pm1.6\%$).
We rigorously attribute this empirical variance to system-level numerical non-determinism, primarily originating from floating-point non-associativity across different batch sizes and non-batch-invariant GPU kernels~\cite{He2025Defeating}. To isolate our algorithmic design from these system-level numerical noises, we conduct a study under a strict ``determinism debug mode'', where we set the dtype to float32 for a higher numerical precision and replace the batch forward with the sequential single-batch forward for FreeDave draft verification to ensure the same GPU kernels (see Appendix~\ref{subsec:determinism} for details). 

As we can observe, all the models achieve 100\% sequence-level and token-level match rates, which proves that \textbf{our claim of losslessness is rigorous} and our code implementation of \freedave~decoding aligns with our theory without any logical errors, and \textbf{the empirical mismatch is attributed to the system-level numerical non-determinism}.
It is worth noting that we primarily focus on the decoding algorithm in this paper, and resolving the non-batch-invariance of GPU kernels is another research problem for future exploration, as we mention in Section~\ref{sec:discussion}. 

\subsection{Ablation Study on Draft Steps}
\label{sec:ablation}

\begin{figure*}[!h]
    \includegraphics[width=1\linewidth]{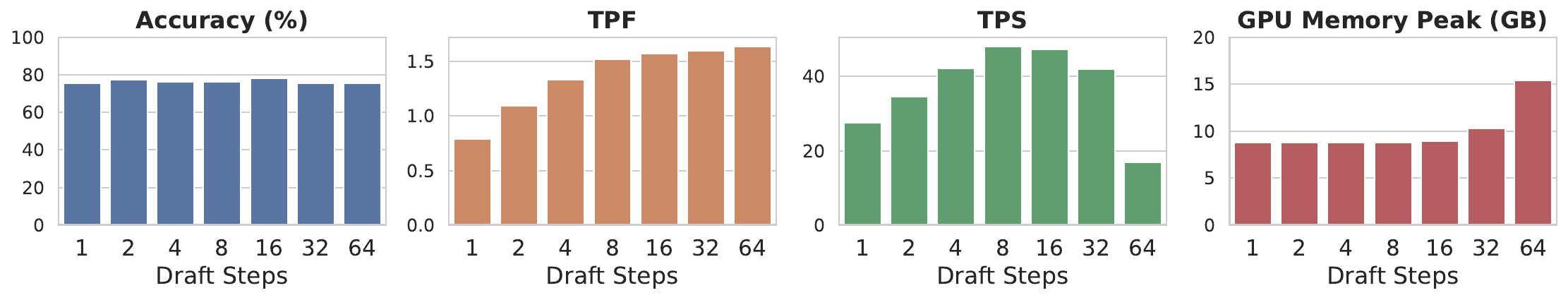}
    \vspace{-10pt}
    \caption{Evaluation of TraDo-4B-Instruct on HumanEval with different draft steps $d$. Specifically, when $d=1$, the decoding algorithm falls back to static decoding. 
    }
    \label{fig:ablation}
\end{figure*}

According to Theorem~\ref{thrm:feasible_path_search} and Lemma~\ref{lemma:optimal_path_guarantee}, a larger $d$ will be more likely to find a shorter feasible path. 
Practically, if $d$ is too large, although we can get fewer decoding steps and a higher TPF, a model forward will take as input a larger tensor, which might take more computation and memory and thus degrade the TPS. 
Thus, in addition to our evaluations on benchmarks, we also investigate the impact of different draft steps $d$ such that we can get the best trade-off between memory overhead and inference efficiency. 
As shown in Figure~\ref{fig:ablation}, we set $d \in \{1, 2, 4, 8, 16, 32, 64\}$, and compare the accuracy, TPS, TPF, and GPU memory peak usage during the inference stage of TraDo-4B-Instruct on HumanEval. As $d$ gets larger, the accuracy remains stable, which aligns with our theoretical analysis of losslessness in Section~\ref{sec:theory}. On the other hand, a larger $d$ means a larger group of draft candidates, which can search shorter paths but will allocate more GPU memory and computation. From Lemma~\ref{lemma:optimal_path_guarantee}, once $d$ is large enough to cover the maximum decision size of the optimal path, we will get the fewest NFE, and from our ablation study, the TPF shows a clear convergence tendency, which perfectly aligns with our theoretical expectation.
However, an indiscriminately large $d$ gets a marginal TPF gain but amplifies the GPU memory footprint (e.g., reaching $\sim15$GB at $d=64$) and prolongs the single-step forward latency, eventually causing a degradation in the wall-clock TPS. Therefore, the selection of $d$ embodies a critical system-level trade-off between theoretical path compression and physical hardware constraints. Empirically, we identify $d=8$ as a universal "sweet spot" for block-attention models, getting an optimal balance that maximizes TPS while incurring a negligible $\sim20$MiB memory overhead compared to static decoding. 

\section{Discussion}
\label{sec:discussion}

We propose \freedave, a novel algorithm performing lossless parallel decoding for dLLMs. By both theoretical analysis and empirical evaluations on math reasoning and code generation benchmarks, our method is proven to bring remarkable inference speedup while maintaining the generation quality, overcoming the challenge of performance degradation faced by threshold-based parallel decoding. 

There are also some future directions remaining. For example, in this paper, we only focus on the offline scenario due to our limited computational resources, and extending our method to online serving with multiple prompt queries on multiple GPUs will be an interesting and critical problem.
Besides, the non-batch-invariant GPU kernels bring numerical fluctuations, and resolving this challenge is also an important AI Infra research problem. 
We leave those problems for future exploration.

\begin{ack}
In this work, Shutong Wu was supported by the Office of the Vice Chancellor for Research and Graduate Education at the University of Wisconsin–Madison with funding from the Wisconsin Alumni Research Foundation and NSF Award DMS-2023239. Jiawei Zhang was supported by the Office of the Vice Chancellor for Research and Graduate Education at the University of Wisconsin–Madison with funding from the Wisconsin Alumni Research Foundation.

The authors declare no competing interests.
\end{ack}


\bibliographystyle{unsrtnat}
\bibliography{neurips_2026}

\newpage
\appendix
\onecolumn

\section{More Algorithm Details}
\subsection{Static Decoding}

For static decoding, the remasking scheduler gives a decision set by 
\begin{equation}
\begin{split}
    \mathcal{I}_{t \rightarrow t-\Delta} &= g_{\text{static}}(x_t, f_\theta(x_t, t), t, t-\Delta)\\
    &= \left\{(a, z) \left| a= \argmax_{i, x_t^i \neq \mask} \left(\max_j f^{i,j}_\theta(x_t, t)\right), z = \argmax_j f^{a,j}_\theta(x_t, t) \right. \right\},
\end{split}
\end{equation}
where $\Delta = 1/L$ is the step size. 
We denote this one-token-per-step greedy decoding strategy as static decoding. Unless specified, in this paper, we focus on the static remasking scheduler $g_{\text{static}}$.

\begin{algorithm}
  \caption{dLLM Static Decoding}
  \label{alg:mdlm_original}
  \begin{algorithmic}[1]
    \Require Model $f_\theta$, remasking scheduler $g_{\text{static}}$, total generation length $L$, total decoding steps $N$, time schedule array $\mathbf{T} = \{t_0, \dots, t_N | 1 = t_0 > t_1 > \dots > t_{N-1} > t_N = 0\}$
    \Ensure The generated sequence $x$.

    \State $x_{t_0} \gets \mask^{1:L}$ 
    \For{$i$ in $0:N-1$}
        \State $\tilde{x}_0 \gets f_\theta(x_{t_i}, t_i)$
        \State $\mathcal{I}_{t_i \rightarrow t_{i+1}} \gets g_{\text{static}}(x_{t_i},\tilde{x}_0, t_i, t_{i+1})$
        \State $x_{t_{i+1}} \gets x_{t_i}$
        \For{each $(a_j, z_j)$ in $\mathcal{I}_{t_i \rightarrow t_{i+1}}$}
            \State $x^{a_j}_{t_{i+1}} \gets z_j$
        \EndFor
    \EndFor
    \State $x \gets x_0$
    \Return $x$
  \end{algorithmic}
\end{algorithm}

\begin{algorithm}[!h]
  \caption{dLLM Threshold-Based Decoding}
  \label{alg:mdlm_threshold}
  \begin{algorithmic}[1]
    \Require Model $f_\theta$, dynamic scheduler $g_{\text{top-k}}$, threshold $\tau$, max length $L$
    \Ensure The generated sequence $x$.

    \State $x \gets \mask^{1:L}$ 
    \While{$\sum_{l=0}^L \mathds{1}(x^l = \mask) > 0$}
        \State $\tilde{x}_0 \gets f_\theta(x_{t_i}, t_i)$
        \State $K_\tau = \min
        \left\{1, \sum_{i,x^i \neq \mask} \mathds{1}(\max_j f^{i,j}_\theta(x) \geq \tau)  \right\}$
        \State $\mathcal{I} \gets g_{\text{top-k}}(x, \tilde{x}_0, K_\tau)$
        \For{each $(a_j, z_j)$ in $\mathcal{I}$}
            \State $x^{a_j} \gets z_j$
        \EndFor
    \EndWhile
    \Return $x$
  \end{algorithmic}
\end{algorithm}

\subsection{Threshold-Based Parallel Decoding}
\label{apdx:threshold}
While static decoding relies on a predefined, fixed time schedule, threshold-based parallel decoding\cite{wu2025fast,yu2025dimple} utilizes a dynamic, data-dependent schedule. At each step, it simultaneously unmasks all tokens whose predicted confidence exceeds a predefined threshold $\tau$.

Because the number of tokens to unmask is dynamically determined, the exact time $t$ for future steps is agnostic beforehand. To formalize this algorithm, we leverage a unique property of absorbing discrete diffusion that the time level $t$ is inherently encoded in the state $x$ itself via the mask ratio. Therefore, we can simplify the model notation from $f_\theta(x_t, t)$ to $f_\theta(x)$, safely omitting the explicit time input. Accordingly, we define this dynamic remasking scheduler that returns a dynamically sized decision set as 
\begin{gather}
    g_{\text{top-k}}(x, f_\theta(x), K_\tau) = \left\{(a, z) \left| a \in \mathop{\text{Top-}K_\tau}\limits_{i, x_t^i \neq \mask} \left(\max_j f^{i,j}_\theta(x)\right), z = \argmax_j f^{a,j}_\theta(x) \right. 
    \right\}\\
    K_\tau = \min
        \left\{1, \sum_{i,x^i \neq \mask} \mathds{1}(\max_j f^{i,j}_\theta(x) \geq \tau)  \right\}
\end{gather}
Here, to ensure decoding progression even when model confidence is generally low, if no token exceeds $\tau$, the scheduler falls back to unmasking the single token with the highest confidence. The standard threshold-based parallel decoding process is detailed in Algorithm~\ref{alg:mdlm_threshold}.

\begin{algorithm}[!h]
  \caption{dLLM Threshold \freedave~Decoding}
  \label{alg:mdlm_freedave++}
  \begin{algorithmic}[1]
    \Require Model $f_\theta$, dynamic scheduler $g_{\text{top-k}}$, threshold $\tau$, max length $L$, draft steps $d$.
    \Ensure The generated sequence $x$.
    
    \State $x \gets \mask^{1:L}$, $\tilde{x}_0 \gets f_\theta(x)$
    \Comment{initialize sequence and its estimated distribution}

    \While{$\sum_{l=0}^L \mathds{1}(x^l = \mask) > 0$}
        \State $\mathbf{X}_{\text{draft}} \gets [~]$, $d_{\text{current}} \gets \min(d, \sum_{l=0}^L \mathds{1}(x^l = \mask))$
        \State $K_\tau = \min
        \left\{1, \sum_{i,x^i \neq \mask} \mathds{1}(\max_j f^{i,j}_\theta(x) \geq \tau)  \right\}$
        \For{$k$ in $1:d_{\text{current}}$}
            \State $\mathcal{I} \gets g_{\text{top-k}}(x, \tilde{x}_0, K_\tau + k - 1)$
            \Comment{$k-1$ more draft tokens in addition to $K_\tau$}
            \State $x_{\text{draft}} \gets x$
            \For{each $(a_j, z_j)$ in $\mathcal{I}$}
            \State $x^{a_j}_{\text{draft}} \gets z_j$
                \Comment{replace masked tokens according to decision}
            \EndFor
            \State $\mathbf{X}_{\text{draft}}\text{.append}(x_{\text{draft}})$
        \EndFor

        \State $\tilde{\mathbf{X}}_{\text{draft}, 0} \gets f_\theta(\mathbf{X}_{\text{draft}})$
            \Comment{all drafts forward in parallel}
        \State $\mathbf{X}_{\text{target}} \gets [~]$
        \For{$k = 1:d_i$}
            \State $K_\tau = \min \left\{1, \sum_{i,x^i \neq \mask} \mathds{1}(\max_j f^{i,j}_\theta(x) \geq \tau)  \right\}$
            \State $\mathcal{I} \gets g_{\text{top-k}}(\mathbf{X}_{\text{draft}}[k], \tilde{\mathbf{X}}_{\text{draft}, 0}[k], K_\tau)$
            \Comment{one more threshold-based step ahead}
            \State $x_{\text{target}} \gets \mathbf{X}_{\text{draft}}[k]$
            \For{each $(a_j, z_j)$ in $\mathcal{I}$}
            \State $x^{a_j}_{\text{target}} \gets z_j$
            \EndFor
            \State $\mathbf{X}_{\text{target}}\text{.append}(x_{\text{target}})$
        \EndFor
        \State $m \gets 0$
        \While{$m < d_{\text{current}} - 1$ }
        \State $m_{\text{ref}} \gets m$
            \For{$k \in \{m+1, \dots,d_{\text{current}}\}$}
                \If {$\mathbf{U}(\mathbf{X}_{\text{draft}}[k]) \subseteq \mathbf{U}(\mathbf{X}_{\text{target}}[m_{\text{ref}}])$}
                    \State $m \gets k$
                \EndIf
            \EndFor
            \If {$m = m_{\text{ref}}$}\
            \Comment{if no more future draft matched, ends verification}
                \State break
            \EndIf
        \EndWhile
        \State $x \gets \mathbf{X}_{\text{draft}}[m + 1], \tilde{x}_0 \gets \tilde{\mathbf{X}}_{\text{draft}, 0}[m + 1]$
    \EndWhile
    \State $x \gets x_0$
    \Return $x$
  \end{algorithmic}
  \vspace{-1pt}
\end{algorithm}

\section{\freedave~Built on Threshold-Based Parallel Decoding}

\label{apdx:threshold_freedave}

As discussed in Section 3.4, \freedave~is orthogonal to the choice of the remasking scheduler and can be seamlessly incorporated with threshold-based parallel decoding to further boost inference efficiency. 
Building upon the notation and the dynamic remasking scheduler $g_{\text{top-k}}(x, \tilde{x}_0, K_\tau)$ established in Appendix~\ref{apdx:threshold}, the Threshold FreeDave algorithm is detailed in Algorithm~\ref{alg:mdlm_freedave++}. 
The core intuition of integrating FreeDave with threshold-based decoding lies in resolving the fundamental conflict between dynamic step sizes and parallel drafting. Because the threshold-based scheduler unmask an unpredictable number of tokens based on real-time confidence, traditional exact-match verification becomes fundamentally incompatible, leading to plummeting acceptance rates.

Algorithm~\ref{alg:mdlm_freedave++} overcomes this through a paradigm shift from strict matching to subset verification. Instead of predicting along fixed time intervals, the algorithm concurrently generates a spectrum of progressive drafts. Crucially, it replaces strict state equality with a relaxed yet rigorous subset criterion. For notation convenience, let $\mathbf{U}(x) = \{i \mid x^i \neq \mathbf{m}, i \in \{1, \dots, L\}\}$ denote the set of unmasked token positions in state $x$. During verification, as long as the tokens unmasked in a draft form a strict subset of the tokens that the threshold heuristic would naturally unmask at the next step (\emph{i.e.}, $\mathbf{U}(\mathbf{X}_{\text{draft}}[k]) \subseteq \mathbf{U}(\mathbf{X}_{\text{target}}[m_{\text{ref}}])$), the draft is deemed logically sound.

By aggressively searching for and accepting the matched draft with the most unmasked tokens, this elegant subset matching flawlessly absorbs the uncertainty of dynamic thresholds. It guarantees that any accepted draft token at one step would have inevitably been selected by the threshold-based scheduler at the next step, thereby achieving extreme parallel acceleration without sacrificing the original generation quality of threshold-based parallel decoding.

\section{Detailed Proof of Lemma~\ref{lemma:optimal_path_guarantee}}
\label{apdx:lemma_proof}

\subsection{Specifications}

Our theoretical setting in Section~\ref{sec:theory} contains two structural specifications: (i) the oracle path is an ordered sequence of transitions along the descending time axis, and feasible merges must preserve this order through a consecutive one-step verification chain; and (ii) the verification frontier is non-retreating along this ordered trajectory. 

\label{apdx:lemma_setting}
\begin{definition}[Ordered One-Step-Verifiable Space]
\label{def:ordered_space}
Let 
\begin{equation}
    x_{t_0} \rightarrow x_{t_1} \rightarrow \cdots \rightarrow x_{t_N}
\end{equation}
be the ordered oracle trajectory. From a trusted state $x_{t_i}$, let $D_{i,r}$ denote the nested draft merging the next $r$ transitions and $T_{i,r}$ its one-step target from the next batched forward. We define $F(i,j)=1$ iff
\begin{equation}
    \label{eq:one_step_verifiable}
    T_{i,r} = D_{i,r+1}, \quad \forall r = 1, \dots, j-i-1,
\end{equation}
\emph{i.e.}, $x_{t_i} \rightarrow \cdots \rightarrow x_{t_j}$ forms a consecutive verification chain. A feasible path is an ordered sequence $0 = a_0 < a_1 < \dots < a_M = N$ with $F(a_r, a_{r+1}) = 1$ for every $r$.
\end{definition}
This ordered specification excludes the infeasible post-mismatch reconvergence: certifying a rejected draft as a later oracle state requires additional sequential forwards from the trusted branch, which have already generated that state and therefore cannot reduce already conducted decoding steps.

Based on $F$, we further denote the farthest one-step-verifiable endpoint and the intrinsic longest merge from $x_{t_i}$ as
\begin{equation}
    \label{eq:farthest_endpoint}
    R(i) = \max\{j : F(i,j) = 1\}, \qquad m(i) = R(i) - i.
\end{equation}
\begin{assumption}[Verification-Frontier Non-Retreat (VFNR)]
\label{assump:vfnr}
\begin{equation}
    \label{eq:vfnr}
    F(i,j) \Rightarrow F(k,j), \quad i < k < j. \tag{VFNR}
\end{equation}
\end{assumption}
Namely, if an oracle interval is verifiable from an earlier state, its remaining successor remains verifiable from any later oracle state within that interval. Equivalently, it can also be interpreted as $R(i+1) \geq R(i)$ or $m(i+1) \geq m(i) - 1$ for $\forall i$. Intuitively, for $i < k < j$, the later oracle state $x_{t_k}$ retains all the unmasked tokens of $x_{t_i}$ and unmasks additional ones, thereby providing strictly more conditioning information for predicting the remaining masked tokens. Hence, if $x_{t_j}$ can already be verified when conditioned on the less informative state $x_{t_i}$, conditioning on the more informative state $x_{t_k}$ should allow the verification to reach at least $x_{t_j}$, \emph{i.e.}, the verification frontier should not retreat as more tokens are unmasked.

\subsection{Proof of Lemma~\ref{lemma:optimal_path_guarantee}}
\label{apdx:lemma_proof_main}
Under these two specifications, let the greedy longest-merge path be $a^\star_0 = 0$, $a^\star_{r+1} = R(a^\star_r)$. Consider any feasible path $0 = b_0 < b_1 < \dots < b_K = N$. We prove Lemma~\ref{lemma:optimal_path_guarantee} by mathematical induction that
\begin{equation}
    \label{eq:greedy_dominance}
    a^\star_r \geq b_r, \quad \forall r.
\end{equation}
Suppose $a^\star_r \geq b_r$. If $a^\star_r \geq b_{r+1}$, then $a^\star_{r+1} \geq b_{r+1}$. Otherwise, $b_r \leq a^\star_r < b_{r+1}$. Since $F(b_r, b_{r+1}) = 1$, VFNR implies $F(a^\star_r, b_{r+1}) = 1$. Because greedy selects the farthest feasible endpoint, we have $a^\star_{r+1} = R(a^\star_r) \geq b_{r+1}$. Thus, $a^\star_r \geq b_r$ implies $a^\star_{r+1} \geq b_{r+1}$, and from $a^\star_0 = b_0$ we inductively prove $a^\star_r \geq b_r, \forall r$. Thus, whenever any feasible path reaches $N$ in $K$ transitions, greedy reaches $N$ in at most $K$ transitions. Hence the greedy longest-merge path is the shortest path in the ordered one-step-verifiable space, and therefore matches the hindsight dynamic programming optimum.

For finite $d$, if $d \geq \max_r (a^\star_{r+1} - a^\star_r)$, no longest merge is truncated, and induction from $a_0 = 0$ yields the optimal path. Thus, $\Vert\mathbf{p}^\star\Vert_\infty$ is a sufficient lower bound for $d$. 

\subsection{Empirical Validity of the VFNR Assumption}
\label{apdx:vfnr_empirical}
To examine how well Assumption~\ref{assump:vfnr} holds in practice, we evaluate Dream-7B-Instruct on 50 MATH500 trajectories. Along each oracle path, we regenerate the drafts at every oracle state and measure the intrinsic longest merge $m(i)$. Across 13902 adjacent state pairs, the condition $m(i+1) \geq m(i) - 1$ holds for 92.78\% of them, indicating that frontier non-retreat is the dominant behavior. 
To assess the impact of the remaining violations, we further construct the complete empirical feasible DAG and compare the greedy longest merge with the hindsight dynamic programming optimum. The greedy longest merge takes 82.30 decoding steps on average, which is only 3.18\% more than the 79.76 steps of the handsight dynamic programming optimum. Therefore, VFNR guarantees the optimality of \freedave~when satisfied, while its occasional violations in practice have only a limited impact on the number of decoding steps.

\section{More Experimental Details}
\label{apdx:exp_detail}

\subsection{Experimental Settings}
Following \citet{wang2025trado}, for the TraDo models, the input prompt is right-padded with mask token $\mask$ such that the block size $B$ divides it, and the total number of generation blocks is $\lceil \frac{(\vert\text{prompt}\vert \bmod B) + \text{max\_gen\_length}}{B} \rceil$.
And for LLaDA-8B-Instruct and Dream-7B-Instruct, the first block starts right after the input prompt without padding. For the attention mask and KV Cache, LLaDA-8B-Instruct and Dream-7B-Instruct use full-sequence bidirectional attention and dual cache~\cite{wu2025fast}, and TraDo-4B-Instruct and TraDo-8B-Instruct use block attention (\emph{i.e.}, intra-block bidirectional and inter-block causal) and prefix cache. 

For the decoding hyperparameters, we use a temperature of 0, and following \citet{wang2025trado}, we use a threshold of 0.95 for LLaDA-8B-Instruct and Dream-7B-Instruct, and 0.9 for the TraDo models.
Besides, an early exit mechanism is applied to all the models. Specifically, when the current block is fully decoded, if all the tokens are $\langle|\textbf{endoftext}|\rangle$, the generation stops, and the current sequence is returned.  
Regarding the attention mask, the TraDo models apply the block-causal mask, which is inter-block causal and intra-block bidirectional, and LLaDA-8B-Instruct and Dream-7B-Instruct apply the full bidirectional mask.

For the inference framework, we build our inference pipeline on plain PyTorch code. We did not use popular engines like vLLM and SGLang, but every experiment run is guaranteed to be under the same test environment. For the evaluation, we incorporate our decoding algorithm into lm-eval-harness \cite{eval-harness}, which is a commonly used evaluation toolkit. All the models are evaluated on a single NVIDIA H200 GPU. Our codebase is publicly available at \url{https://github.com/cychomatica/FreeDave}.

\subsection{Detailed Results of Benchmark Evaluations}
\label{subsec:benchmark_details}
In Table~\ref{tab:llada_detail}–\ref{tab:trado8b_detail}, we provide the detailed numbers of benchmark evaluations. For math reasoning benchmarks, we choose MATH500 \cite{hendrycks2021measuring} and GSM8K \cite{cobbe2021training}.
For code generation benchmarks, we choose MBPP \cite{austin2021program} and HumanEval \cite{chen2021evaluating}. 
It is worth noting that the TPF of static decoding for TraDo models is around 0.8 instead of 1 due to a separate model forward for cache refresh after every block is fully decoded. Since the block size is 4, there will be 5 model forwards every 4 decoded tokens.
\vspace{10pt}
\begin{table}[!h]
    \centering
    \small
    \begin{tabularx}{\textwidth}{clLLL} 
    \toprule
    \textbf{Benchmark} & \textbf{Decoding Strategy} & \textbf{Acc (\%)~$\uparrow$} & \textbf{TPF~$\uparrow$} & \textbf{TPS~$\uparrow$} \\    \midrule
    \multirow{4}{*}{\makecell{MATH500\\(4-shot)}} & Static & 37.40 & 1.00 & 37.05 \\
     & Static \freedave~(Ours) & 36.80 \down{0.60} & 2.45 \speedup{2.45} & 74.19 \speedup{2.00} \\
     & Threshold & 37.40 \up{0.00} & 2.61 \speedup{2.61} & 85.89 \speedup{2.32} \\
     & Threshold \freedave~(Ours) & 36.80 \down{0.60} & 4.40 \speedup{4.40} & 119.47 \speedup{3.22} \\
    \cmidrule{1-5}
    \multirow{4}{*}{\makecell{GSM8K\\(5-shot)}} & Static & 78.77 & 1.00 & 34.23 \\
     & Static \freedave~(Ours) & 78.62 \down{0.15} & 2.50 \speedup{2.50} & 68.52 \speedup{2.00} \\
     & Threshold & 78.85 \up{0.08} & 2.61 \speedup{2.61} & 77.47 \speedup{2.26} \\
     & Threshold \freedave~(Ours) & 78.09 \down{0.68} & 4.39 \speedup{4.39} & 105.94 \speedup{3.09} \\
    \cmidrule{1-5}
    \multirow{4}{*}{\makecell{MBPP\\(3-shot)}} & Static & 36.00 & 1.00 & 36.16 \\
     & Static \freedave~(Ours) & 36.20 \up{0.20} & 2.58 \speedup{2.58} & 73.34 \speedup{2.03} \\
     & Threshold & 36.00 \up{0.00} & 2.74 \speedup{2.74} & 82.48 \speedup{2.28} \\
     & Threshold \freedave~(Ours) & 36.20 \up{0.20} & 4.81 \speedup{4.81} & 116.60 \speedup{3.22} \\
    \cmidrule{1-5}
    \multirow{4}{*}{\makecell{HumanEval\\(0-shot)}} & Static & 35.98 & 1.00 & 41.58 \\
     & Static \freedave~(Ours) & 35.98 \up{0.00} & 2.45 \speedup{2.45} & 84.18 \speedup{2.02} \\
     & Threshold & 35.98 \up{0.00} & 3.33 \speedup{3.33} & 120.97 \speedup{2.91} \\
     & Threshold \freedave~(Ours) & 35.98 \up{0.00} & 5.48 \speedup{5.48} & 163.75 \speedup{3.94} \\
    \bottomrule
    \end{tabularx}
    \caption{LLaDA-8B-Instruct results on math reasoning and code generation benchmarks.}
    \label{tab:llada_detail}
\end{table}

\begin{table}[!h]
    \centering
    \small
    \begin{tabularx}{\textwidth}{clLLL} 
    \toprule
    \textbf{Benchmark} & \textbf{Decoding Strategy} & \textbf{Acc (\%)~$\uparrow$} & \textbf{TPF~$\uparrow$} & \textbf{TPS~$\uparrow$} \\    \midrule
    \multirow{4}{*}{\makecell{MATH500\\(4-shot)}} & Static & 44.80 & 1.00 & 62.51 \\
     & Static \freedave~(Ours) & 46.20 \up{1.40} & 2.45 \speedup{2.45} & 115.16 \speedup{1.84} \\
     & Threshold & 44.60 \down{0.20} & 2.35 \speedup{2.35} & 125.59 \speedup{2.01} \\
     & Threshold \freedave~(Ours) & 46.40 \up{1.60} & 3.97 \speedup{3.97} & 163.34 \speedup{2.61} \\
    \cmidrule{1-5}
    \multirow{4}{*}{\makecell{GSM8K\\(5-shot)}} & Static & 79.91 & 1.00 & 58.86 \\
     & Static \freedave~(Ours) & 79.23 \down{0.68} & 2.49 \speedup{2.49} & 107.58 \speedup{1.83} \\
     & Threshold & 79.83 \down{0.08} & 2.52 \speedup{2.52} & 121.40 \speedup{2.06} \\
     & Threshold \freedave~(Ours) & 78.77 \down{1.14} & 4.10 \speedup{4.10} & 150.35 \speedup{2.55} \\
    \cmidrule{1-5}
    \multirow{4}{*}{\makecell{MBPP\\(3-shot)}} & Static & 52.80 & 1.00 & 62.66 \\
     & Static \freedave~(Ours) & 53.80 \up{1.00} & 2.67 \speedup{2.67} & 123.06 \speedup{1.96} \\
     & Threshold & 53.00 \up{0.20} & 2.60 \speedup{2.60} & 136.96 \speedup{2.19} \\
     & Threshold \freedave~(Ours) & 54.00 \up{1.20} & 4.27 \speedup{4.27} & 171.56 \speedup{2.74} \\
    \cmidrule{1-5}
    \multirow{4}{*}{\makecell{HumanEval\\(0-shot)}} & Static & 54.88 & 1.00 & 68.74 \\
     & Static \freedave~(Ours) & 56.10 \up{1.22} & 2.60 \speedup{2.60} & 133.95 \speedup{1.95} \\
     & Threshold & 54.88 \up{0.00} & 2.40 \speedup{2.40} & 146.19 \speedup{2.13} \\
     & Threshold \freedave~(Ours) & 56.10 \up{1.22} & 4.03 \speedup{4.03} & 185.08 \speedup{2.69} \\
    \bottomrule
    \end{tabularx}
    \caption{Dream-v0-Instruct-7B results on math reasoning and code generation benchmarks.}
    \label{tab:dream_detail}
\end{table}

\begin{table}[!h]
    \centering
    \small
    \begin{tabularx}{\textwidth}{clLLL} 
    \toprule
    \textbf{Benchmark} & \textbf{Decoding Strategy} & \textbf{Acc (\%)~$\uparrow$} & \textbf{TPF~$\uparrow$} & \textbf{TPS~$\uparrow$} \\    \midrule
    \multirow{4}{*}{\makecell{MATH500\\(4-shot)}} & Static & 64.80 & 0.80 & 28.63 \\
     & Static \freedave~(Ours) & 66.40 \up{1.60} & 1.68 \speedup{2.10} & 54.69 \speedup{1.91} \\
     & Threshold & 65.60 \up{0.80} & 1.51 \speedup{1.89} & 53.71 \speedup{1.88} \\
     & Threshold \freedave~(Ours) & 65.40 \up{0.60} & 1.94 \speedup{2.42} & 61.65 \speedup{2.15} \\
    \cmidrule{1-5}
    \multirow{4}{*}{\makecell{GSM8K\\(5-shot)}} & Static & 90.67 & 0.79 & 27.69 \\
     & Static \freedave~(Ours) & 90.52 \down{0.15} & 1.64 \speedup{2.08} & 52.97 \speedup{1.91} \\
     & Threshold & 90.37 \down{0.30} & 1.42 \speedup{1.80} & 50.14 \speedup{1.81} \\
     & Threshold \freedave~(Ours) & 89.76 \down{0.91} & 1.92 \speedup{2.43} & 59.57 \speedup{2.15} \\
    \cmidrule{1-5}
    \multirow{4}{*}{\makecell{MBPP\\(3-shot)}} & Static & 60.40 & 0.79 & 27.01 \\
     & Static \freedave~(Ours) & 60.00 \down{0.40} & 1.52 \speedup{1.92} & 47.17 \speedup{1.75} \\
     & Threshold & 56.00 \down{4.40} & 1.21 \speedup{1.53} & 42.39 \speedup{1.57} \\
     & Threshold \freedave~(Ours) & 57.20 \down{3.20} & 1.80 \speedup{2.28} & 55.02 \speedup{2.04} \\
    \cmidrule{1-5}
    \multirow{4}{*}{\makecell{HumanEval\\(0-shot)}} & Static & 75.61 & 0.79 & 27.50 \\
     & Static \freedave~(Ours) & 76.22 \up{0.61} & 1.52 \speedup{1.92} & 48.17 \speedup{1.75} \\
     & Threshold & 76.22 \up{0.61} & 1.14 \speedup{1.44} & 39.55 \speedup{1.44} \\
     & Threshold \freedave~(Ours) & 77.44 \up{1.83} & 1.72 \speedup{2.18} & 52.88 \speedup{1.92} \\
    \bottomrule
    \end{tabularx}
    \caption{TraDo-4B-Instruct results on math reasoning and code generation benchmarks.}
    \label{tab:trado4b_detail}
\end{table}

\begin{table}[!h]
    \centering
    \small
    \begin{tabularx}{\textwidth}{clLLL} 
    \toprule
    \textbf{Benchmark} & \textbf{Decoding Strategy} & \textbf{Acc (\%)~$\uparrow$} & \textbf{TPF~$\uparrow$} & \textbf{TPS~$\uparrow$} \\    \midrule
    \multirow{4}{*}{\makecell{MATH500\\(4-shot)}} & Static & 69.20 & 0.80 & 26.29 \\
     & Static \freedave~(Ours) & 69.40 \up{0.20} & 1.64 \speedup{2.05} & 51.68 \speedup{1.97} \\
     & Threshold & 69.40 \up{0.20} & 1.41 \speedup{1.76} & 46.43 \speedup{1.77} \\
     & Threshold \freedave~(Ours) & 69.20 \up{0.00} & 1.90 \speedup{2.37} & 58.67 \speedup{2.23} \\
    \cmidrule{1-5}
    \multirow{4}{*}{\makecell{GSM8K\\(5-shot)}} & Static & 93.03 & 0.80 & 25.45 \\
     & Static \freedave~(Ours) & 92.87 \down{0.16} & 1.62 \speedup{2.02} & 50.47 \speedup{1.98} \\
     & Threshold & 92.49 \down{0.54} & 1.37 \speedup{1.71} & 44.41 \speedup{1.74} \\
     & Threshold \freedave~(Ours) & 92.57 \down{0.46} & 1.90 \speedup{2.37} & 60.25 \speedup{2.37} \\
    \cmidrule{1-5}
    \multirow{4}{*}{\makecell{MBPP\\(3-shot)}} & Static & 65.40 & 0.79 & 24.38 \\
     & Static \freedave~(Ours) & 65.80 \up{0.40} & 1.48 \speedup{1.87} & 45.30 \speedup{1.86} \\
     & Threshold & 63.80 \down{1.60} & 1.16 \speedup{1.47} & 37.48 \speedup{1.54} \\
     & Threshold \freedave~(Ours) & 64.20 \down{1.20} & 1.70 \speedup{2.15} & 52.31 \speedup{2.15} \\
    \cmidrule{1-5}
    \multirow{4}{*}{\makecell{HumanEval\\(0-shot)}} & Static & 83.54 & 0.79 & 25.35 \\
     & Static \freedave~(Ours) & 83.54 \up{0.00} & 1.52 \speedup{1.92} & 45.94 \speedup{1.81} \\
     & Threshold & 82.93 \down{0.61} & 1.15 \speedup{1.46} & 35.20 \speedup{1.39} \\
     & Threshold \freedave~(Ours) & 81.71 \down{1.83} & 1.69 \speedup{2.14} & 51.01 \speedup{2.01} \\
    \bottomrule
    \end{tabularx}
    \caption{TraDo-8B-Instruct results on math reasoning and code generation benchmarks.}
    \label{tab:trado8b_detail}
\end{table}

\subsection{Details of Non-determinism Attribution Experiments}
\label{subsec:determinism}
We rigorously attribute this empirical variance to system-level numerical non-determinism, primarily originating from floating-point non-associativity across different batch sizes and non-batch-invariant GPU kernels~\cite{He2025Defeating}. To isolate our algorithmic design from these hardware-level execution noises, we conduct an ablation study under a strict ``determinism debug mode'' with the following factors controlled:
\begin{itemize}
\item Set the data type of model parameters and tensors to float32 for a higher numerical precision. 
\item Replace the forward operation of concatenated \freedave~draft branches with an individual forward operation for each draft branch, such that the model computes tensors with the same size and the same GPU kernels are used by both static decoding and \freedave~decoding.  
\end{itemize}

Under the determinism debug mode, for different models on HumanEval, we calculated the sequence-level and token-level exact match rate. Specifically, the sequence-level match rate is defined as
\begin{equation}
    R_{\text{sequence}}(\mathbf{X}, \mathbf{Y}) = \frac{\sum_{i=1}^N \mathds{1} (\mathbf{X}_i = \mathbf{Y}_i)}{N},
\end{equation}
and the token-level match rate is defined as
\begin{equation}
    R_{\text{token}}(\mathbf{X}, \mathbf{Y}) = \frac{\sum_{i=1}^N \vert \text{LongestSharedPrefix}(\mathbf{X}_i, \mathbf{Y}_i) \vert}{\sum_{i=1}^N \max \left( \vert \mathbf{X}_i \vert, \vert \mathbf{Y}_i \vert \right)},
\end{equation}
where $\mathbf{X}_i$ and $\mathbf{Y}_i$ are the outputs static and FreeDave decoding for the $i$-th input, respectively, and $N$ is the total number of inputs. The all-$100\%$ match rate results are shown in Table~\ref{tab:determinism_debug}. 

\subsection{Ablation Details}
\begin{figure}
    \centering
    \includegraphics[width=0.99\linewidth]{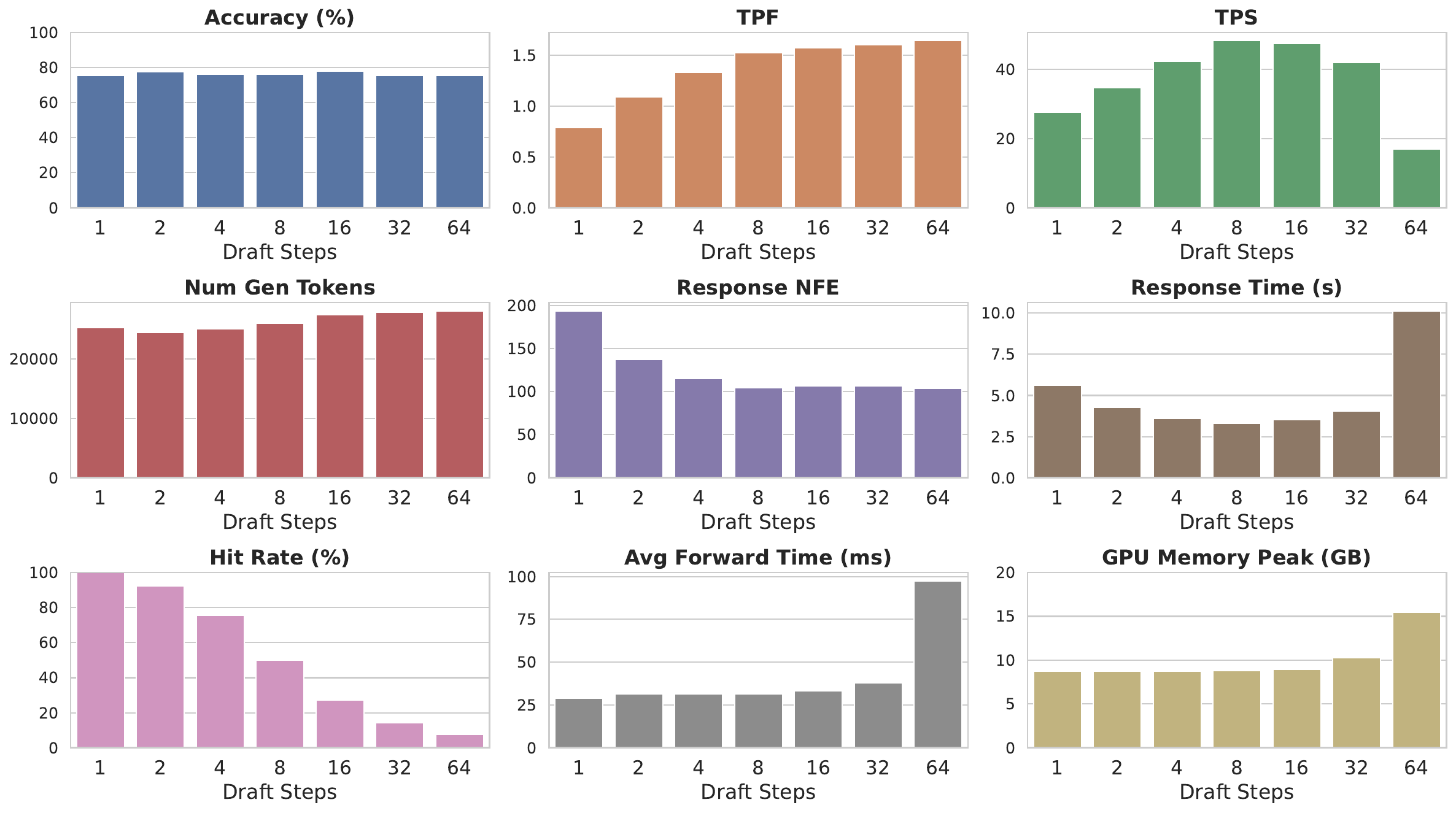}
    \caption{Ablation details}
    \label{fig:ablation_detail}
\end{figure}

In addition to our evaluations on benchmarks, we also investigate the impact of different draft steps $d$ such that we can get the best trade-off between memory overhead and inference efficiency. As shown in Figure~\ref{fig:ablation_detail}, we set $d \in \{1, 2, 4, 8, 16, 32, 64\}$, and compare the accuracy, TPS, TPF, number of generated tokens, response NFE, response time, hit rate, average forward pass time, and GPU memory peak usage during the inference stage of TraDo-4B-Instruct on HumanEval. Specifically, hit rate (\emph{i.e.}, acceptance rate) is calculated by the number of accepted tokens divided by the number of draft tokens.

\subsection{Additional Evaluations on Q\&A Benchmarks}
\label{apdx:qa_eval}
In addition to math reasoning and code generation, we also apply our method to different models and compare with other baselines on Q\&A benchmarks, including GPQA\cite{rein2024gpqa} and TruthfulQA\cite{lin2022truthfulqa}. Specifically, for TruthfulQA, we use \verb|rougeL_acc| as the accuracy metric. The detailed results are shown in Table~\ref{tab:qa_detail}.

\begin{table}[!h]
  \caption{Different models on Q\&A benchmarks.}
  \label{tab:qa_detail}
  \centering
  \setlength{\tabcolsep}{4pt}
  \footnotesize
  \begin{tabularx}{\textwidth}{cllCCC} 
    \toprule
    \textbf{Benchmark} & \textbf{Model} & \textbf{Decoding Strategy} & \textbf{Acc (\%)~$\uparrow$} & \textbf{TPF~$\uparrow$} & \textbf{TPS~$\uparrow$} \\    \midrule
    \multirow{16}{*}{\makecell{GPQA\\(0-shot)}} & \multirow{4}{*}{LLaDA-8B-Instruct} &  Static & 23.44 & 1.00 & 43.01 \\
    & & Static \freedave~(Ours) & 21.43 \down{2.01} & 2.37 \speedup{2.37} & 84.50 \speedup{1.96} \\
    & & Threshold & 23.21 \down{0.23} & 2.13 \speedup{2.13} & 85.16 \speedup{1.98} \\
    & & Threshold \freedave~(Ours) & 21.65 \down{1.79} & 3.67 \speedup{3.67} & 123.06 \speedup{2.86} \\
    \cmidrule{2-6}
    & \multirow{4}{*}{Dream-7B-Instruct} & Static & 27.23 & 1.00 & 69.53 \\
    & & Static \freedave~(Ours) & 26.79 \down{0.44} & 2.44 \speedup{2.44} & 130.34 \speedup{1.87} \\
    & & Threshold & 27.23 \up{0.00} & 2.29 \speedup{2.29} & 144.85 \speedup{2.08} \\
    & & Threshold \freedave~(Ours) & 26.79 \down{0.44} & 3.87 \speedup{3.87} & 188.48 \speedup{2.71} \\
    \cmidrule{2-6}
    & \multirow{4}{*}{TraDo-4B-Instruct} & Static & 19.42 & 0.80 & 25.92 \\
    & & Static \freedave~(Ours) & 21.43 \up{2.01} & 1.53 \speedup{1.91} & 47.89 \speedup{1.85} \\
    & & Threshold & 21.88 \up{2.46} & 1.22 \speedup{1.52} & 39.41 \speedup{1.52} \\
    & & Threshold \freedave~(Ours) & 21.88 \up{2.46} & 1.70 \speedup{2.12} & 52.68 \speedup{2.03} \\
    \cmidrule{2-6}
    & \multirow{4}{*}{TraDo-8B-Instruct} & Static & 34.82 & 0.80 & 25.98 \\
    & & Static \freedave~(Ours) & 34.82 \up{0.00} & 1.44 \speedup{1.80} & 44.48 \speedup{1.71} \\
    & & Threshold & 35.04 \up{0.22} & 1.09 \speedup{1.36} & 35.22 \speedup{1.36} \\
    & & Threshold \freedave~(Ours) & 35.04 \up{0.22} & 1.57 \speedup{1.96} & 48.17 \speedup{1.85} \\
    \cmidrule{1-6}
    \multirow{16}{*}{\makecell{TruthfulQA\\(0-shot)}} & \multirow{4}{*}{LLaDA-8B-Instruct} & Static & 38.56 & 1.00 & 42.76 \\
    & & Static \freedave~(Ours) & 38.31 \down{0.25} & 2.38 \speedup{2.38} & 83.91 \speedup{1.96} \\
    & & Threshold & 38.56 \up{0.00} & 1.96 \speedup{1.96} & 78.13 \speedup{1.83} \\
    & & Threshold \freedave~(Ours) & 38.43 \down{0.13} & 3.52 \speedup{3.52} & 115.00 \speedup{2.69} \\
    \cmidrule{2-6}
    & \multirow{4}{*}{Dream-7B-Instruct} & Static & 52.63 & 1.00 & 71.46 \\
    & & Static \freedave~(Ours) & 52.88 \up{0.25} & 2.87 \speedup{2.87} & 147.87 \speedup{2.07} \\
    & & Threshold & 52.51 \down{0.12} & 2.21 \speedup{2.21} & 143.13 \speedup{2.00} \\
    & & Threshold \freedave~(Ours) & 52.88 \up{0.25} & 4.29 \speedup{4.29} & 202.56 \speedup{2.83} \\
    \cmidrule{2-6}
    & \multirow{4}{*}{TraDo-4B-Instruct} & Static & 49.57 & 0.78 & 24.61 \\
    & & Static \freedave~(Ours) & 49.57 \up{0.00} & 1.41 \speedup{1.81} & 42.69 \speedup{1.73} \\
    & & Threshold & 48.71 \down{0.86} & 0.95 \speedup{1.22} & 29.88 \speedup{1.21} \\
    & & Threshold \freedave~(Ours) & 49.08 \down{0.49} & 1.52 \speedup{1.95} & 45.97 \speedup{1.87} \\
    \cmidrule{2-6}
    & \multirow{4}{*}{TraDo-8B-Instruct} & Static & 55.69 & 0.78 & 24.75 \\
    & & Static \freedave~(Ours) & 55.20 \down{0.49} & 1.42 \speedup{1.82} & 41.93 \speedup{1.69} \\
    & & Threshold & 55.69 \up{0.00} & 0.95 \speedup{1.22} & 29.42 \speedup{1.19} \\
    & & Threshold \freedave~(Ours) & 55.81 \up{0.12} & 1.54 \speedup{1.97} & 45.12 \speedup{1.82} \\
    \bottomrule
  \end{tabularx}
\end{table}

\section{Memory Cost Analysis}
\label{apdx:memory_cost}

\subsection{Theoretical Memory Complexity}
\label{apdx:memory_complexity}
Let $b$ denote the inference batch size, $B$ the block size, $d$ the number of draft steps, and $L$ the total sequence length (\emph{i.e.}, prompt plus generated tokens). Compared to static decoding, which evaluates the $B$ tokens of the current block in each forward pass, \freedave~evaluates the current block together with the $d-1$ additional draft branches, \emph{i.e.}, $(d-1)B$ additional tokens. Under FlashAttention~\cite{dao2022flashattention}, which avoids materializing the $L \times L$ attention matrix and reduces the attention memory complexity from quadratic to linear in the sequence length, the peak memory costs of static decoding and \freedave~can be expressed as
\begin{equation}
    \label{eq:mem_static}
    M_{\text{Static}} = M_\theta + M_0 + \mathcal{O}\left(bL\,m_{\text{tok}}\right),
\end{equation}
\begin{equation}
    \label{eq:mem_freedave}
    M_{\text{\freedave}} = M_\theta + M_0 + \mathcal{O}\left(b\left(L + (d-1)B\right) m_{\text{tok}}\right),
\end{equation}
where $M_\theta$ denotes the memory for loading the model weights, $M_0$ the runtime overheads (\emph{e.g.}, CUDA context and allocator workspace), and $m_{\text{tok}}$ the working-memory footprint per token (\emph{e.g.}, KV cache and activations). Consequently, the additional memory introduced by \freedave~is
\begin{equation}
    \label{eq:mem_overhead}
    \Delta M = M_{\text{\freedave}} - M_{\text{Static}} = \mathcal{O}\left(b(d-1)B\, m_{\text{tok}}\right).
\end{equation}
Thus, the overhead of \freedave~scales linearly with the batch size $b$, the block size $B$, and the number of draft steps $d$, which is consistent with the growth of peak memory \emph{w.r.t.} $d$ observed in our ablation study in Figure~\ref{fig:ablation_detail}. Importantly, when $b$, $B$, and $d$ are fixed, $\Delta M$ is independent of the sequence length $L$, and the relative overhead $\Delta M / (M_{\text{Static}} - M_\theta - M_0) = \mathcal{O}\left((d-1)B / L\right)$ diminishes as $L$ grows. 

\subsection{Empirical Validation with Different Generation Lengths}
\label{apdx:memory_empirical}
\begin{figure}
    \centering
    \includegraphics[width=0.99\linewidth]{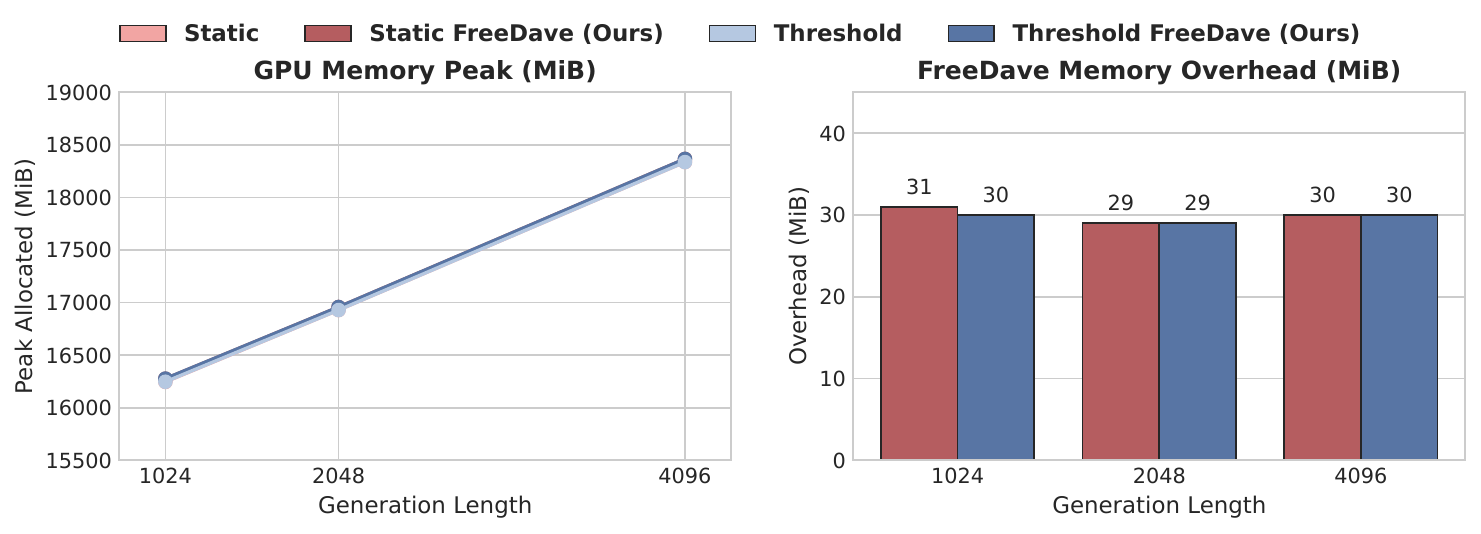}
    \caption{GPU peak memory and \freedave~memory overhead for Dream-7B-Instruct on MATH500 under different generation lengths. }
    \label{fig:memory_vs_length}
\end{figure}

Empirically, we evaluate Dream-7B-Instruct on MATH500 with different generation lengths in $\{1024, 2048, 4096\}$, using the same block size $B=32$ and draft steps $d=4$ as in Section~\ref{sec:exp:settings}. Specifically, the memory for loading the model weights under bf16 is $M_\theta = 14569$ MiB. The detailed results are shown in Table~\ref{tab:memory_long_gen}, and the peak allocated memory is further visualized in Figure~\ref{fig:memory_vs_length}.

\begin{table}
  \caption{Results of Dream-7B-Instruct on MATH500 under different generation lengths.}
  \label{tab:memory_long_gen}
  \centering
  \setlength{\tabcolsep}{4pt}
  \footnotesize
  \begin{tabularx}{\textwidth}{clCCCCC}
    \toprule
    \multirow{2}{*}{\makecell{\textbf{Generation}\\\textbf{Length}}} & \multirow{2}{*}{\textbf{Decoding Strategy}} & \multirow{2}{*}{\textbf{Acc (\%)~$\uparrow$}} & \multirow{2}{*}{\textbf{TPF~$\uparrow$}} & \multirow{2}{*}{\textbf{TPS~$\uparrow$}} & \multicolumn{2}{c}{\textbf{GPU Peak Memory (MiB)}} \\
    \cmidrule{6-7}
    & & & & & \textbf{Allocated} & \textbf{Reserved} \\
\midrule
    \multirow{4}{*}{1024} & Static & 44.80 & 1.00 & 62.51 & 16245 & 16746 \\
     & Static \freedave~(Ours) & 46.20 & 2.45 & 115.16 & 16276 & 16752 \\
     & Threshold & 44.60 & 2.35 & 125.59 & 16247 & 16746 \\
     & Threshold \freedave~(Ours) & 46.40 & 3.97 & 163.34 & 16277 & 16752 \\
    \cmidrule{1-7}
    \multirow{4}{*}{2048} & Static & 44.80 & 1.00 & 50.21 & 16929 & 17624 \\
     & Static \freedave~(Ours) & 45.80 & 2.40 & 87.61 & 16958 & 17630 \\
     & Threshold & 44.80 & 2.53 & 99.84 & 16929 & 17624 \\
     & Threshold \freedave~(Ours) & 45.40 & 4.16 & 123.87 & 16958 & 17630 \\
    \cmidrule{1-7}
    \multirow{4}{*}{4096} & Static & 44.60 & 1.00 & 35.86 & 18337 & 19552 \\
     & Static \freedave~(Ours) & 44.60 & 2.36 & 57.93 & 18367 & 19558 \\
     & Threshold & 44.60 & 3.12 & 72.10 & 18337 & 19552 \\
     & Threshold \freedave~(Ours) & 44.20 & 5.03 & 83.76 & 18367 & 19558 \\
    \bottomrule
  \end{tabularx}
\end{table}

As shown in Figure~\ref{fig:memory_vs_length}, the peak memory of all the decoding strategies grows linearly with the generation length, which matches the $\mathcal{O}(bL\,m_{\text{tok}})$ term in Eq.~\eqref{eq:mem_static} and Eq.~\eqref{eq:mem_freedave}. Meanwhile, the additional memory of \freedave~over its base decoding strategy remains nearly constant ($\sim$30 MiB) across all the generation lengths, which is consistent with Eq.~\eqref{eq:mem_overhead}, where $\Delta M$ depends only on $b$, $B$, and $d$ rather than $L$. 

\newpage
\section{Details of Block Diffusion Extensions}

\subsection{Block Tree Attention Details}
\label{apdx:block_tree_attn}
\begin{figure}[!h]
    \centering
    \includegraphics[width=0.99\linewidth]{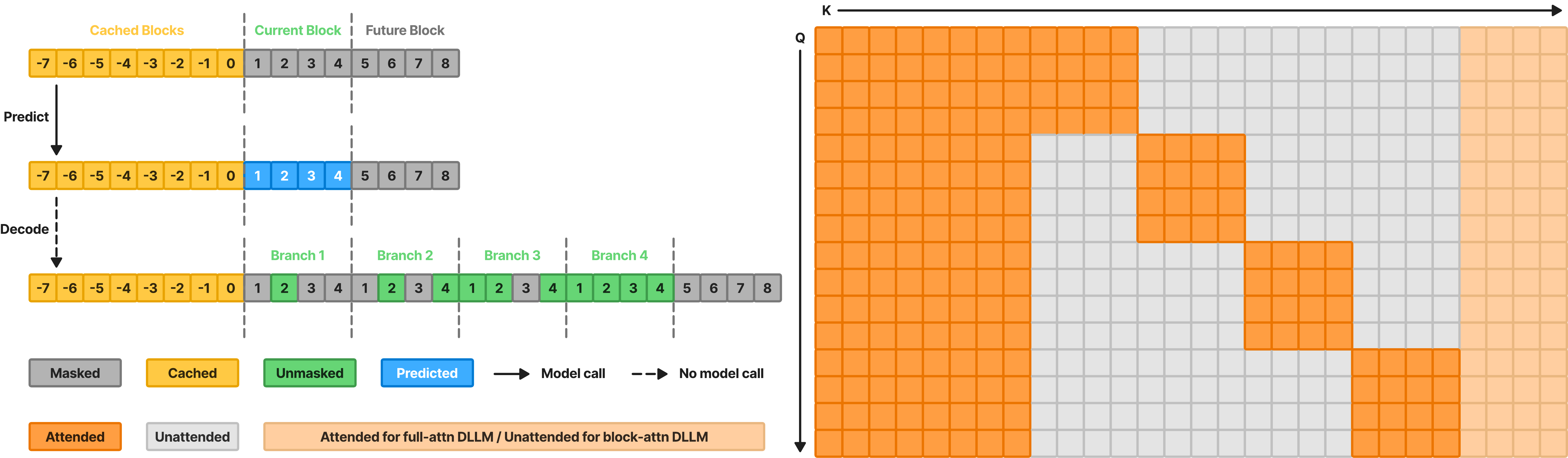}
    \caption{\freedave~with block tree attention tailored for block diffusion.}
    \label{fig:block_tree_attn}
\end{figure}

For block diffusion, instead of expanding draft candidates along the batch dimension, which incurs substantial memory overhead, we compactly pack all draft candidates into a single sequence and employ a block-level tree attention topology, as illustrated in Figure~\ref{fig:block_tree_attn}. This structure prevents redundant KV cache allocation for shared prefixes across different draft candidates and reduces the memory footprint during parallel verification. 

\subsection{Eager Mode Details and the Equivalence to Static Decoding Variant}
\label{apdx:eager_mode_details}

\begin{figure}[!h]
    \centering
    \includegraphics[width=1\linewidth]{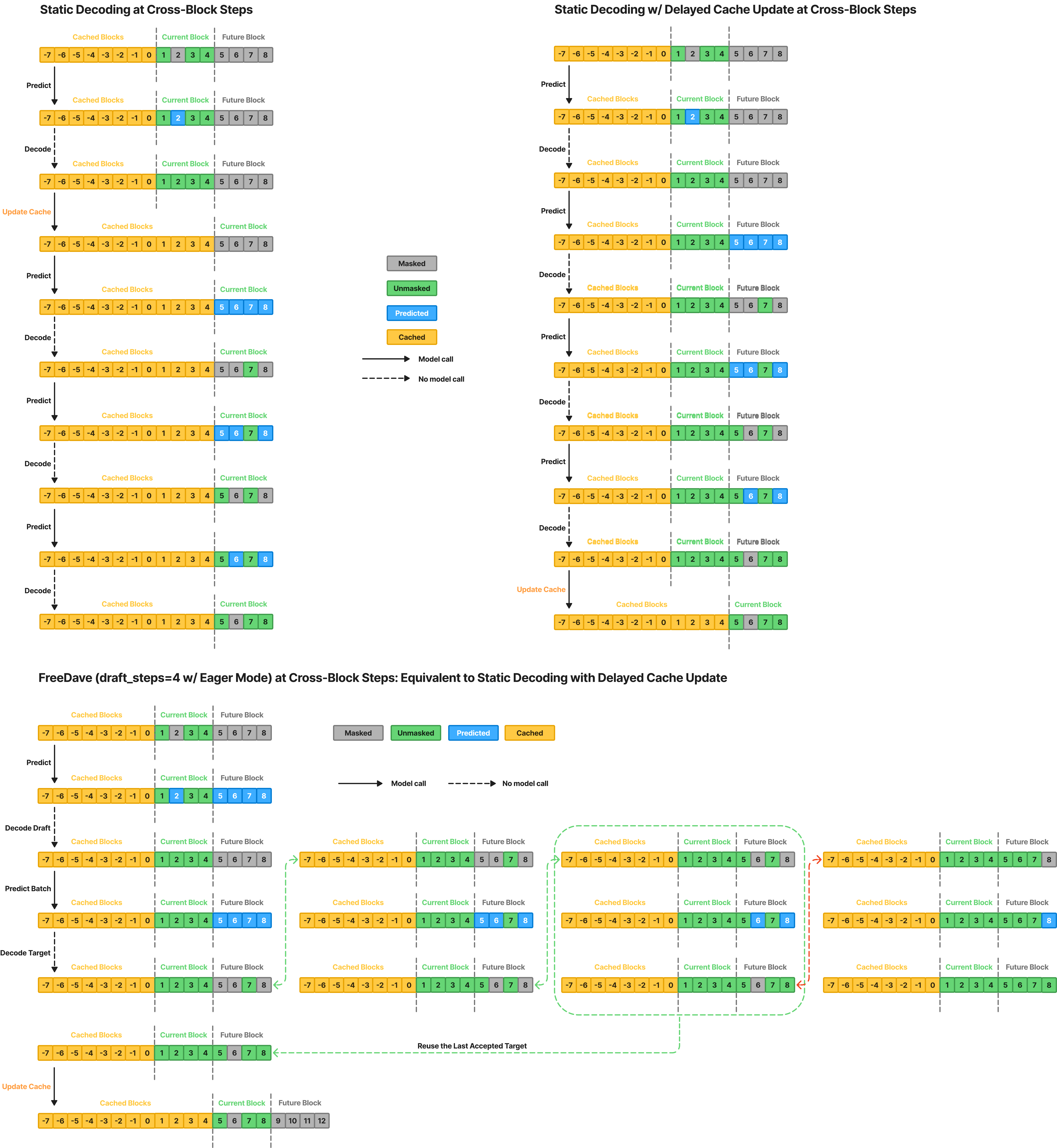}
    \caption{The equivalence between \freedave~Eager Mode and the variant of static decoding with delayed cache update.}
    \label{fig:eager_mode_detail}
    \vspace{-5pt}
\end{figure}

Eager Mode permits pre-drafting and pre-verifying tokens across future blocks while maintaining a strict left-to-right acceptance priority. This not only eliminates boundary-induced truncations but also explicitly enables scaling the draft budget $d$ beyond the block size itself. 
More importantly, as demonstrated in Figure~\ref{fig:eager_mode_detail}, Eager Mode is structurally equivalent to a static decoding variant with delayed cache updates, thereby strictly preserving the generation quality while unlocking significantly higher decoding efficiency. 
In the example of Figure~\ref{fig:eager_mode_detail}, for standard static decoding (upper left), the tokens $x^5$, $x^7$, and $x^8$ are generated using cached KV of $x^{-7:4}$. When the cache update is delayed (upper right), $x^5$, $x^7$, and $x^8$ are generated using cached KV of $x^{-7:0}$ and computed KV of $x^{1:4}$, which aligns with \freedave~with Eager Mode enabled. The only difference is the cached and computed KV of $x^{1:4}$, where the latter should be more informative and at least as good as the former.



\end{document}